\documentclass{article}

\usepackage[preprint]{preprint}

\usepackage[utf8]{inputenc} 
\usepackage[T1]{fontenc}    
\usepackage{hyperref}       
\usepackage{url}            
\usepackage{booktabs}       
\usepackage{amsfonts}       
\usepackage{nicefrac}       
\usepackage{microtype}      
\usepackage{xcolor}         
\usepackage{graphicx}
\usepackage{caption}
\usepackage{float}
\usepackage{amsmath} 
\usepackage{amsfonts} 
\usepackage{mathtools} 
\usepackage{array}
\usepackage{algorithm}
\usepackage{algpseudocode}
\usepackage{placeins}

\title{KunPeng: A Global Ocean Environmental Model}

\author{
	\begin{tabular}{c}
		\begin{tabular}{@{}c@{\hspace{2em}}c@{\hspace{2em}}c@{}}
			Yi Zhao      & Jiaqi Li     & Haitao Xia   \\
			Tianjiao Zhang\textsuperscript{*} & Zerong Zeng & Tianyu Ren \\ 
			Yucheng Zhang & Chao Zhu     & Shengtong Xu \\
			& Hongchun Yuan\textsuperscript{†} & \\
		\end{tabular}
		\\
		Shanghai Ocean University \\
		\vspace{0.3em}
		\texttt{\{m230901642, 2352204, 1925223, m240701902,} \\  
		\texttt{m240751972, 2452334, m240701901, 2452801\}@st.shou.edu.cn} \\
		\texttt{tjzhang@shou.edu.cn\textsuperscript{*}, hcyuan@shou.edu.cn\textsuperscript{†}} \\ 
		\vspace{0.5em}
	\end{tabular}
}

\begin{document}

\maketitle

\begin{abstract}
 Ocean environmental forecasting is a core component in studying global climate  change, disaster warning, and fishery prediction.  Traditional numerical models  are limited by parameterization schemes and computational resources, making  it challenging to accurately characterize multi-scale ocean dynamical processes.
 Although deep learning technology offers a new paradigm for ocean modeling, existing dedicated ocean models still face challenges such as observational data sparsity, insufficient physical consistency, and multi-scale feature coupling.
 Inspired by the similarity of the atmosphere-ocean physical coupling mechanism, this study innovatively migrates meteorological large-model techniques to the ocean  domain, constructing the KunPeng global ocean environmental prediction model.
 Aimed at the discontinuous characteristics of marine space, we propose a terrain-adaptive mask constraint mechanism to mitigate effectively training divergence  caused by abrupt gradients at land-sea boundaries.
 To fully integrate far-, medium-, and close-range marine features, a longitude-cyclic deformable convolution network (LC-DCN) is employed to enhance the dynamic receptive field, achieving refined modeling of multi-scale oceanic characteristics.A Deformable Convolution-enhanced Multi-Step Prediction module (DC-MTP) is employed to strengthen temporal dependency feature extraction capabilities.
 Experimental results demonstrate that this model achieves an average ACC of 0.80 in 15-day global predictions at 0.25$^\circ$ resolution, outperforming comparative models by 0.01-0.08. The average mean squared error (MSE) is 0.41 (representing a 5\%–31\% reduction) and the average mean absolute error (MAE) is 0.44 (0.6\%–21\% reduction) compared to other models. Significant improvements are particularly observed in sea surface parameter prediction, deep-sea region characterization, and current velocity field forecasting.Through a horizontal comparison of the applicability of operators at different scales in the marine domain, this study reveals that local operators significantly outperform global operators under slow-varying oceanic processes, demonstrating the effectiveness of dynamic feature pyramid representations in predicting marine physical parameters.
 Finally, this study not only validates the feasibility of cross-domain model transfer but also establishes the first global ocean model benchmarking framework, laying theoretical and technical foundations for high-resolution coupled ocean-atmosphere models. 
 The code for our proposed model and baseline models is publicly available at \href{https://github.com/kbdsbx/kunpeng}{https://github.com/kbdsbx/kunpeng}.%
\end{abstract}

\section{Introduction}
\label{introduction}

In the northern ocean there is a fish, called the {\bf Kun}, I do not know how many thousand li in size. This Kun changes into a bird, called the {\bf Peng}. Its back is I do not know how many thousand li in  breadth. (Zhuangzi, \emph{Free and Easy Wandering}, 4th Century BC). In ancient times, people envisioned  the tumultuous waves of the sea as colossal fish, while perceiving the boundless expanse of clouds as massive birds soaring through the heavens.This anthropomorphic symbolism reflects humanity’s enduring ambition to decipher the behavioral patterns of natural forces, harness their latent potential, and transform these primordial energies into sustainable utilities rather than catastrophic events - a transcendental aspiration persisting through millennia. Contemporary advances in deep learning are now emerging as a transformative catalyst toward realizing this archetypal vision.

With global warming, frequent extreme weather events, and marine pollution, accurate ocean environmental forecasting has become a frontier topic in international research field. Traditional numerical models face bottlenecks in coupling multi-scale oceanic processes due to simplified parameterizations and computational limitations. In recent years, the breakthrough advancements in artificial intelligence technologies, spearheaded by deep learning, have ushered in a new paradigm for marine environmental modeling. Large neural network-based models, integrating multi-source remote sensing, argo buoy arrays, and reanalysis data, demonstrate strong nonlinear mapping capabilities for sea surface temperature, ocean current movement, and typhoon trajectories. Advances in graph convolutional networks and Transformers architecture algorithm enable capturing atmosphere-ocean interactions while addressing effectively prediction biases in data-sparse deep-sea regions. These intelligent models and their derivative applications, applied in disaster warning, fishery assessment, and polar ice monitoring, signify a paradigm shift toward data-driven modeling synergized with oceanographic principles.

In recent years, the application of large-scale deep learning models in ocean environmental modeling  has garnered increasing attention, yet remains in its nascent stage compared to advancements in the meteorological domain. Pioneering efforts, exemplified by AI-GOMS~\cite{RN20} and the XiHe model~\cite{RN21}, have demonstrated end-to-end predictions for critical oceanic  variables—such as sea surface temperature, salinity, and current velocity—using deep neural networks. However, existing dedicated ocean models continue to face significant challenges in generalization  capability and physical consistency, due to the sparsity of marine observational data and the strongly nonlinear nature of oceanic dynamic processes. Meanwhile, the meteorological domain has witnessed  breakthroughs in large-model technologies. For instance, FourCastNet~\cite{RN22} employs  Fourier neural operators to deliver global atmospheric predictions at 0.25$^\circ$ spatial resolution, achieving  inference speeds 80,000 times faster than conventional numerical models.  GraphCast~\cite{RN23}, built upon graph neural networks, constructs implicit representations of three-dimensional atmospheric dynamics, attaining optimal accuracy for 90\% of its predefined 1,380 prediction targets. The Pangu model~\cite{RN24}, adopting 3D Earth-specific Transformer (3DEST), has become the first to surpass the operational forecasting system of the European Centre for Medium-Range Weather Forecasts (ECMWF) in terms of 10-day prediction accuracy. These models, by integrating physical prior knowledge with data-driven approaches, demonstrate marked advantages in interpretability and computational efficiency, thereby providing prior feasibility validation for cross-domain applications  bridging oceanic and atmospheric systems.

Ocean forecasting faces dual challenges compared to the weather prediction. Spatially, the representation of marine environmental parameters exhibit discontinuous characteristics and multi-scale  temporal coupling. In the spatial dimension, ocean tensors are segmented by continental boundaries  and influenced by submarine topographic gradients (for instance the East Pacific Rise with topographic variations exceeding 3,000 meters), forming highly dynamic masked regions. For instance, under a Mercator projection with a 0.25$^\circ$×0.25$^\circ$ sea surface grid, approximately 38\% of grid cells require specialized handling due to land coverage. Temporally, oceanic processes evolve slowly, with the processes of ocean currents and thermal changes operating at longer temporal scales~\cite{RN25}. Additionally, existing technical approach derived from atmospheric models show limitations in oceanic contexts. While adaptive Fourier neural operators and Transformer-based global attention mechanisms effectively capture rapidly varying features like atmospheric jets, their uniform allocation of computational resources leads to inefficient modeling of slow-varying marine processes. This phenomenon—termed "overfitting of long-range dependencies and under-parameterization of short-range correlations" reveals the adaptability limitations of purely global operators in resolving  the multi-scale physics inherent to ocean systems.

To address the aforementioned challenges, this study innovatively adapts meteorological foundation models to the ocean domain. Based on the physical similarity in atmosphere-ocean coupling mechanisms we construct a cross-domain evaluation framework through feature space remapping and physics-informed fine-tuning. To tackle the spatial discontinuity and multi-scale coupling inherent in marine systems, we propose the KunPeng Global Ocean Model with three key innovations: A topography-adaptive masked loss constraint resolving training divergence from land-sea  boundary gradients (cyclic padding applied along coastlines); Longitude-cyclic deformable convolution enabling mesoscale eddy modeling (50–200 km) via dynamic receptive field adjustment; A deformable convolution-enhanced multi-step predictor capturing temporally lagged feature dependencies. To enable systematic cross-model evaluation of ocean models in real-world scenarios, this study pioneers a global ocean model intercomparison framework. Through multi-dimensional, multi-scale, and multi-region analyses encompassing temporal dynamics, spatial patterns, and metric performance, the framework comprehensively assesses model capabilities. Furthermore, cross-domain validation experiments are introduced to examine non-metric criteria including prediction smoothness and indicator correlation strength. These advancements provide both theoretical foundations for improving oceanic physical field forecasting accuracy and technical prerequisites for developing next-generation high-resolution global ocean-atmosphere coupled models.

\section{Results}
\subsection{Multiple Models 15-Day Prediction Comparison}

The multi-model comparative analysis demonstrates that in 15-day medium-range forecasting evaluations, the KunPeng model demonstrates significant advantages over baseline models such as Pangu~\cite{RN24}, AI-GOMS~\cite{RN20}, and FourCastNet~\cite{RN22}, while showing a clear tendency in comparison with the GraphCast~\cite{RN23} baseline model. Specifically, in temperature (\(T\)) forecasting performance, KunPeng is slightly inferior to GraphCast in 60\% of metrics, primarily concentrated in the 10--200\,m ocean layer, with 20\% of metrics remaining on par (Remaining on par means that in the 15-day forecast, the values are higher/lower than those of the comparison model for 6 to 9 days, and the situation is reversed for the remaining days.) and 20\% outperforming GraphCast. In salinity (\(S\)) forecasting performance, KunPeng outperforms GraphCast in 70\% of metrics, with 7\% being comparable and 23\% inferior, primarily in MAE metrics for the 30-150\,m ocean layer. In three-dimensional velocity field (\(U/V\)) reconstruction, the KunPeng model surpasses the GraphCast baseline across all depth layers, achieving best performance in all comparative model evaluations. Notably, the AI-GOMS model shows a slight advantage in short-term forecasting (3 days) of surface velocity fields (1\,m/5\,m/10\,m), but as the forecast horizon extends beyond 5 days, its cumulative error growth rate significantly exceeds that of KunPeng and GraphCast. Additionally, in key evaluation metrics T1-ACC and S1-MSE, the KunPeng model achieves continuous error reduction through a dynamic error correction mechanism, optimizing forecasting accuracy for predictions beyond 5 days. Partial metric comparisons are detailed in Figure~\ref{fig:metric_comparison}, with complete metric comparison data provided in Appendix~\ref{appendix:C}.

\begin{figure}[htbp]
	\centering
	\includegraphics[width=\linewidth]{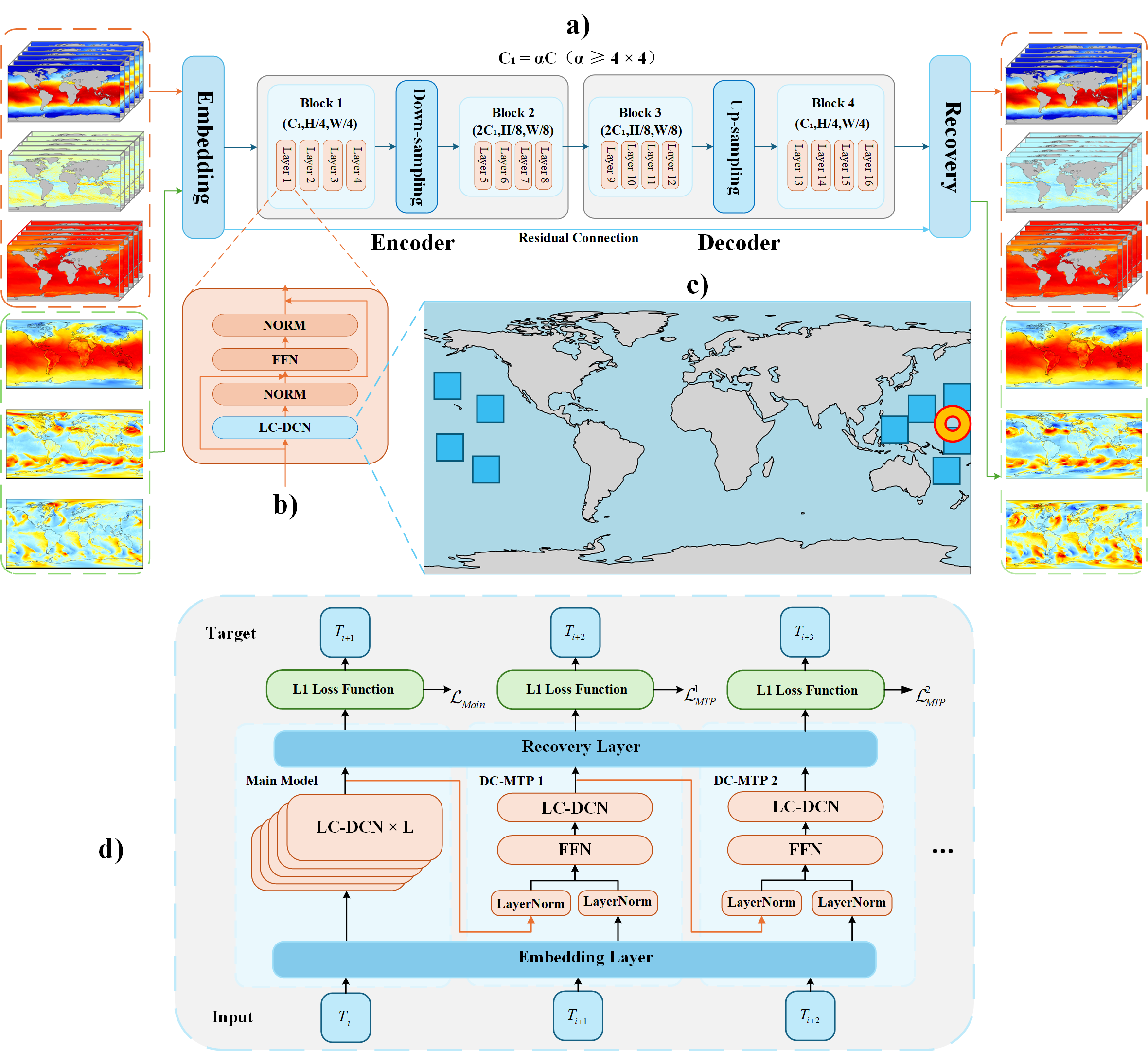}  
	\caption{Architecture of the KunPeng Ocean Large Model. \textbf{a)} is basic model structure. The input consists of three-dimensional ocean environmental parameters, two-dimensional ocean environmental parameters, meteorological environmental parameters at time T, and the static field, which is stacked along the coastal layer channels to form a ($C$,$H$,$W$)-dimensional tensor. After passing through the Embedding layer, a ($C_1$,$H/4$,$W/4$)-dimensional hidden-layer feature is obtained. The feature mixing part is composed of an Encoder and a Decoder, each of which contains two operator blocks, and each block consists of four layers with the same structure. In the Encoder, the two blocks are connected by a down-sampling layer. Through a convolution with a stride of 2, the hidden layer is sampled into a ($2C_1$,$H/8$,$W/8$)-dimensional tensor. The up-sampling layer of the Decoder uses a transposed convolution with the same stride to perform the opposite operation. Finally, the Recovery layer is used to restore the block and reconstruct the ($C_2$,$H$,$W$)-dimensional atmospheric/oceanic physical field at time T+1. Generally, $C_2 \leq C_{1}$ to remove the static fields that do not need to be predicted, such as the ocean layer depth. To ensure numerical stability, there is a residual connection between the Embedding and Recovery layers. \textbf{b)} is Single-layer Structure in the Operator Block. Here, Norm represents layer normalization, FFN is the feed-forward neural network, which is mainly used for feature channel mixing, and LC-DCN is the longitude-cyclic deformable convolution, which is mainly used for feature space mixing. Each mixer has a residual link.  \textbf{c)} is Schematic Diagram of the Longitude-cyclic Deformable Convolution. The double-circled points represent the pixel positions for the convolution calculation, and the squares represent the positions of the convolution kernel units after being adjusted by the deformation offset. Some units of the same convolution kernel can cross the left-right boundaries to perform the longitude-cyclic convolution operation. \textbf{d)} is Schematic Diagram of the DC-MTP(Deformable Convolution-enhanced Multi-step Prediction). The primary model's loss function $L_{main}$ and the auxiliary prediction model's loss function $L^{i}_{MTP}$ both employ Masked Dimension-Weighted L1 Loss (MLL1). The auxiliary model is only used during training to improve the main model's ability to capture temporal features and will be removed during inference.}
	\label{fig:pig1}
\end{figure}

\begin{figure}[htbp]
	\centering
	\includegraphics[width=0.8\linewidth]{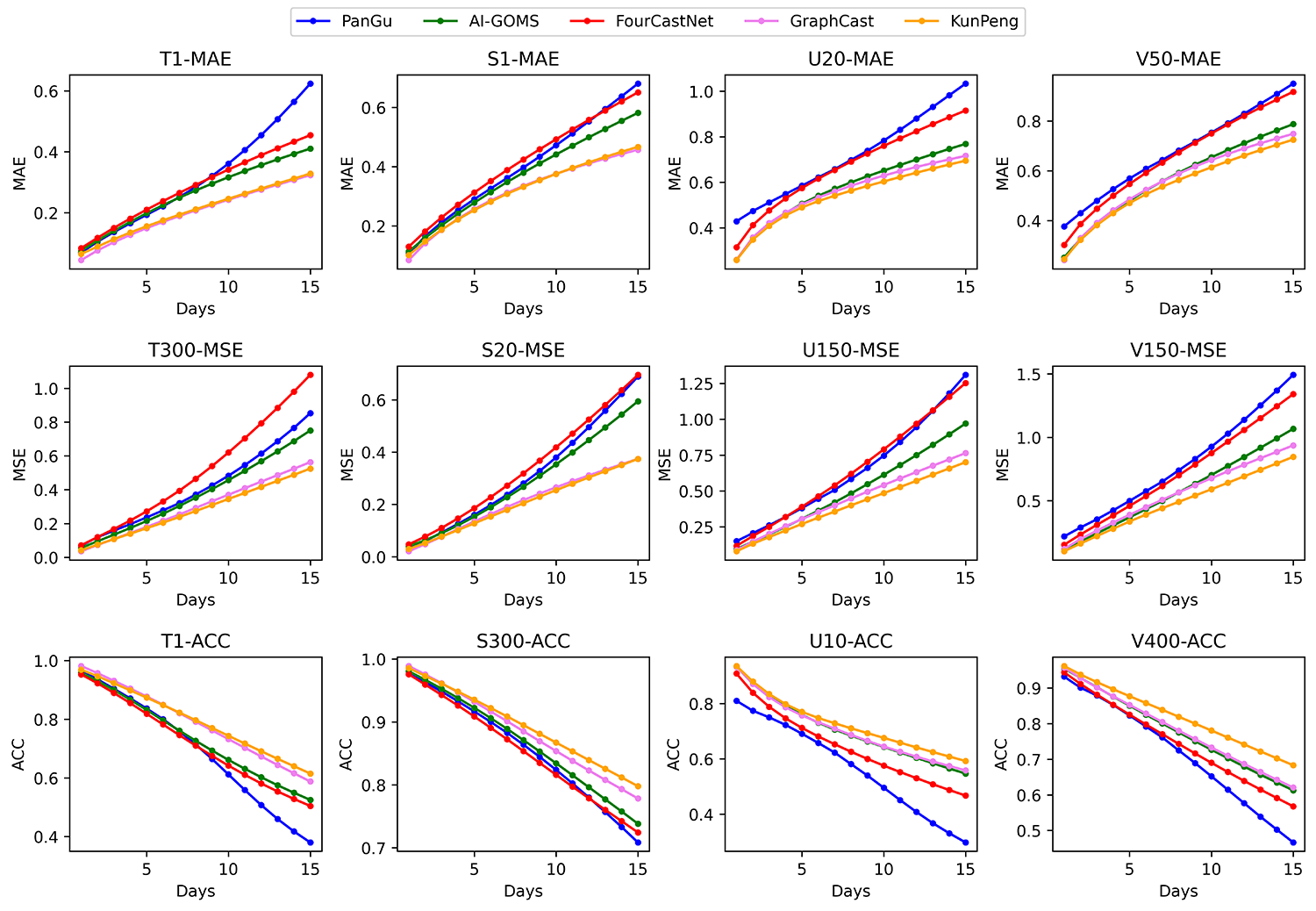} 
	\caption{Comparison of selected model metrics. Here, \(T\) denotes temperature, \(S\) salinity, \(U\) meridional current velocity, and \(V\) zonal current velocity. Numerical labels indicate sea layer depth. Metrics were derived from daily 15-day forecast results spanning \texttt{2021-01-01T00:00:00Z UTC} to \texttt{2021-12-16T00:00:00Z UTC}. Lower values of MAE and MSE indicate better performance, with a theoretical minimum of 0, while higher ACC values signify improved accuracy, approaching a theoretical optimum of 1.}
	\label{fig:metric_comparison}
\end{figure}

\begin{figure}[htbp]
	\centering
	\includegraphics[width=0.8\linewidth]{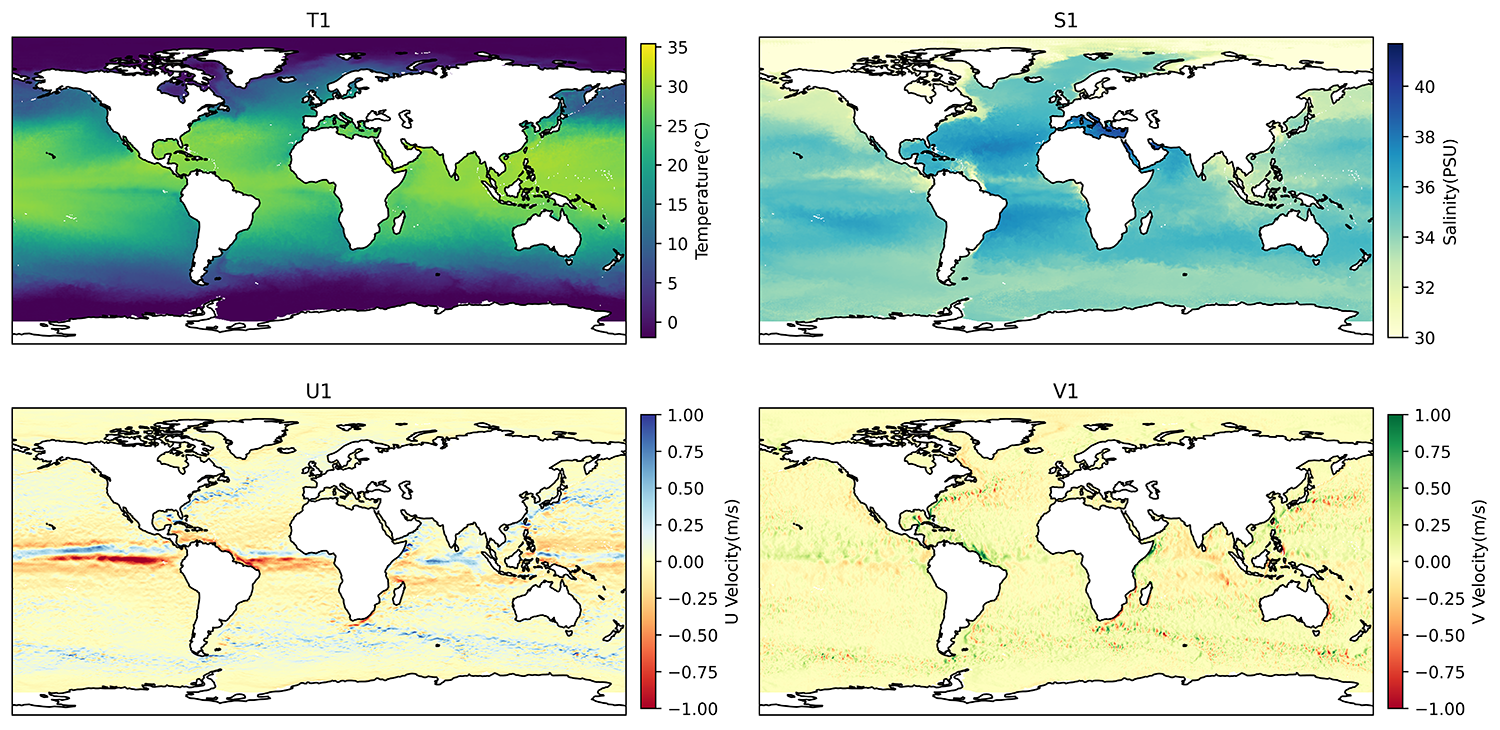}
	\caption{This is a prediction map of the global ocean area at \texttt{2021-07-16T00:00:00Z UTC}. This prediction was obtained through 15 cycles of data using the data at \texttt{2021-07-01T00:00:00Z UTC} as the initial value. Among them, $T_1$, $S_1$, $U_1$, and $V_1$ represent the sea temperature at a depth of 1 meter, salinity, meridional current velocity, and zonal current velocity, respectively.}
	\label{fig:prediction_maps}
\end{figure}

Figure~\ref{fig:prediction_maps} shows global ocean prediction maps based on reanalysis data from \texttt{2021-07-01T00:00:00Z UTC}, with 15 iterative predictions yielding results for \texttt{2021-07-16T00:00:00Z UTC}, with the sea surface physical fields at 1 m depth being selected for analysis. Table~\ref{tab:model_performance} further quantifies the models’ comprehensive performance in daily prediction tasks throughout 2021. By averaging metrics over the 15-day prediction period, the results indicate that the KunPeng model demonstrates significant advantages in key evaluation metrics. The 15-day forecast achieved an average Anomaly Correlation Coefficient (ACC) of 0.80, surpassing comparative models by an improvement margin of 0.01--0.08. The mean values of Mean Squared Error (MSE) and Mean Absolute Error (MAE) stood at 0.40 and 0.44, respectively, demonstrating relative reductions of 5\%--31\% for MSE and 0.6\%--21\% for MAE. All metrics positioned the model as the top-performing among all evaluated models.

\begin{table}[htbp]
	\centering
	\caption{Mean values of 15-day prediction metrics}
	\label{tab:model_performance}
	\begin{tabular}{l *{3}{>{\centering\arraybackslash}p{3cm}}} 
		\toprule
		Model & MAE & MSE & ACC \\
		\midrule
		Pangu & 0.55 & 0.59 & 0.72 \\
		AI-GOMS & 0.50 & 0.50 & 0.77 \\
		FourCastNet & 0.56 & 0.59 & 0.74 \\
		GraphCast & 0.44 & 0.43 & 0.79 \\
		\textbf{KunPeng(ours)} & \textbf{0.44} & \textbf{0.40} & \textbf{0.80} \\
		\bottomrule
	\end{tabular}
\end{table}

\subsection{Regional Sea Comparison}

\begin{figure}[htbp]
	\centering
	\includegraphics[width=0.8\linewidth]{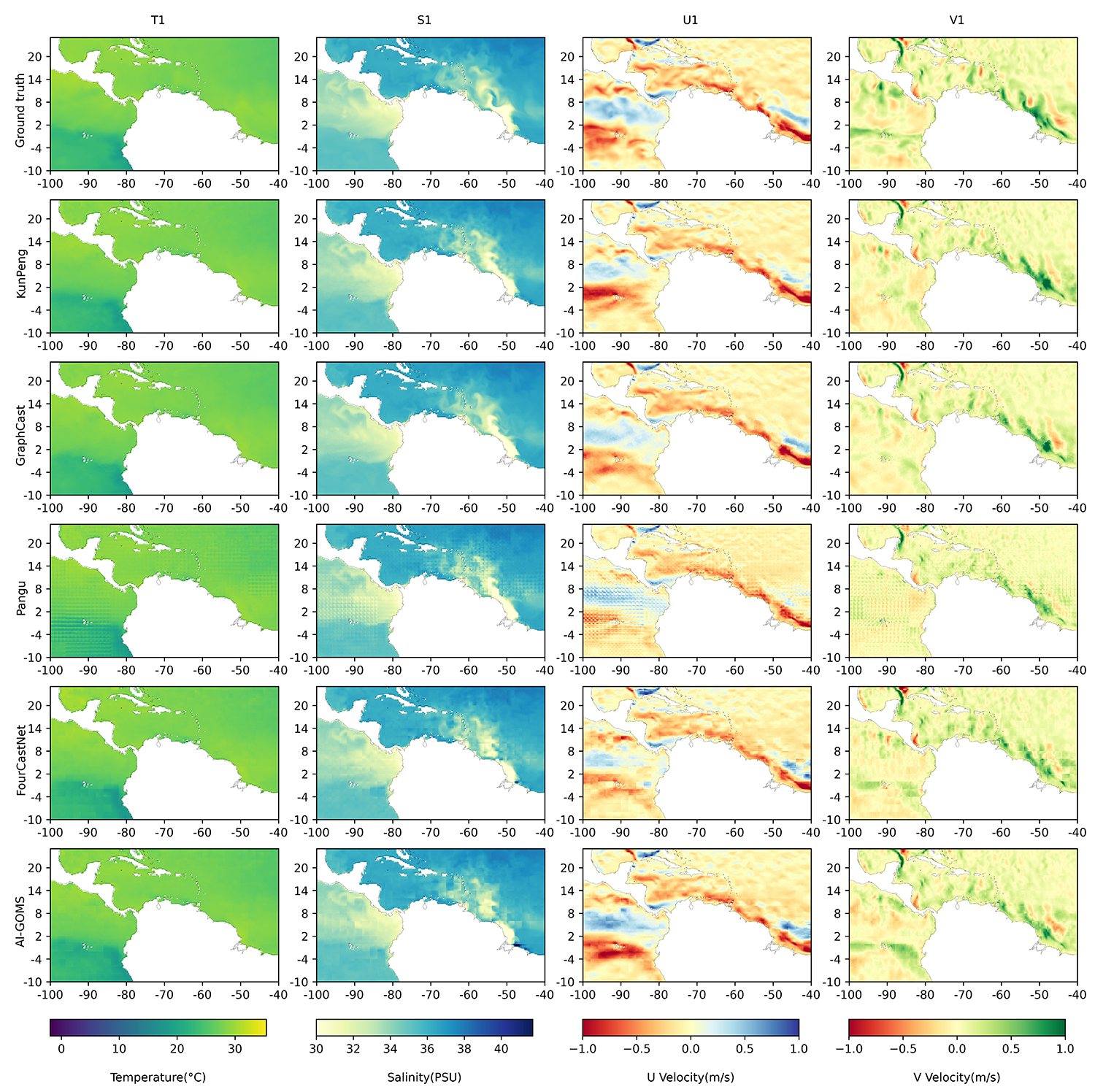}
	\caption{This is a prediction map of the region between $10^\circ$S and $25^\circ$N in latitude between $100^\circ$W and $40^\circ$W in longitude at \texttt{2021-07-11T00:00:00Z UTC}. This prediction was derived from 10 cycles of data based on the data at \texttt{2021-07-01T00:00:00Z UTC}. Among them, $T1$, $S1$, $U1$, and $V1$ represent the sea temperature at a depth of 1 m, salinity, meridional current velocity, and zonal current velocity, respectively.}
	\label{fig:forecast_map}
\end{figure}

This study selects the 1-meter water layer in the tropical Eastern Pacific to the Atlantic Ocean (10$^\circ$S--25$^\circ$N, 100$^\circ$W--40$^\circ$W) from \texttt{2021-07-01T00:00:00Z UTC}, to predict conditions on \texttt{2021-07-11T00:00:00Z UTC}, as the research area (Figure~\ref{fig:forecast_map}) to evaluate the performance differences of multiple models in simulating mesoscale ocean processes. The experimental results show that, due to model architecture limitations, AI-GOMS, FourCastNet, and Pangu models exhibit significant error accumulation effects in continuous predictions, presenting noticeable blocky segmentation at a 0.25$^\circ$ spatial resolution. In contrast, the KunPeng model demonstrates higher stability at the same resolution through an effective iterative prediction correction mechanism, significantly improving the representation of slow-varying ocean processes.

Notably, the study area shows prediction differences at the boundary of the Amazon River estuary (49$^\circ$W, 0$^\circ$): the salinity prediction errors of all models on the Western Pacific side (100$^\circ$W-50$^\circ$W) are significantly lower than those on the Atlantic side (50$^\circ$W-40$^\circ$W). Mechanism analysis reveals that, although a mask loss function is used to isolate land interference, the Amazon River's annual freshwater discharge of \(1.75 \times 10^5 \, \text{m}^3/\text{s}\) still indirectly couples with adjacent ocean regions, leading to systematic biases in salinity prediction. This highlights the need for future model development to strengthen the coupling modeling of land-sea boundary processes.

\section{Conclusions and Discussions}
\label{conclusion}
\subsection{Adaptation of Computer Vision Operators in Ocean Environment Prediction}

Computer vision and multimodal large language models, as core research directions in the large-scale pre-trained model technology system, continue to drive the innovative development of intelligent computing paradigms. At the data representation level, the typical input of computer vision systems is three-dimensional tensor data structures (width $\times$ height $\times$ channel), which shares the same form as the gridded meteorological/oceanic physical field data used in climate numerical prediction. The main difference lies in the latter's multi-dimensional spatiotemporal resolution and high channel complexity. Based on this, applying the neural network architecture in the field of image processing to the fields of the atmosphere and ocean essentially constitutes the domain transfer of the feature extraction paradigm.

However, it should be noted that the geometric differences between the Earth's ellipsoid and Euclidean space lead to systematic errors in traditional planar projection methods. Limited by the processing paradigm of existing computational architectures for regular tensor data, the mainstream methods still employ the Mercator projection method to generate regular tensors, which inevitably introduces geometric distortion errors, necessitating reverse correction of prediction results. Particularly in oceanic physical field modeling, the coupling effect between fluid continuity characteristics and data projection distortion, such as the segmentation of the International Date Line (180$^\circ$ longitude) on the spatial continuity characteristics of the Pacific region, makes it difficult for local computational units to effectively capture adjacent physical field correlations.

Notably, the slow-varying nature of the ocean system (caused by the high viscosity coefficient of water, which prolongs feature interaction time) imposes special requirements on model architecture design: traditional global attention mechanisms, when computing feature correlations, treat distant low-correlation regions with equal weights, leading to computational resource redundancy. GraphCast innovatively adopts a spherical graph neural network architecture. By constructing a spherical subdivision topological graph to maintain consistency with geographical features, and by retaining the edge connections of all subdivision levels to construct multi-scale meteorological interactions.

However, this approach still has two key flaws: first, the recursive grid partitioning algorithm leads to uneven distribution of adjacent edges among grid nodes, with higher adjacency degrees near the top grid and underrepresentation of nodes with lower adjacency degrees; second, the coupling mechanism of spatial feature fusion and channel feature transformation (implemented using a fully connected feedforward network for feature transformation and message passing) significantly increases model complexity, resulting in a peak memory consumption of 47,531 MiB during the inference phase, three times higher than other models (approximately 15,000 MiB), severely limiting its deployment feasibility on consumer-grade GPU devices.

KunPeng was designed with these limitations in mind and has made improvements. As a local operator, convolution is more suitable for the slow-varying characteristics of the ocean. Deformable convolution can automatically enhance local sampling density in high-gradient regions (such as land-sea boundaries) while expanding the receptive field in smooth regions (such as open ocean). By stacking multiple layers of deformable convolution units, the bottom network captures small-scale turbulent features, and the top network aggregates large-scale circulation patterns, forming a feature pyramid structure from local to global. Furthermore, we propose longitude-cyclic deformable convolution, which enables the convolution kernel to seamlessly capture feature propagation across the International Date Line while eliminating the pseudo-circulation closure phenomenon caused by traditional boundary padding. Combined with the dynamic sensing capability of deformable convolution, the improved network can enhance local sampling density at abrupt regions like land-sea boundaries while maintaining the global continuity of large-scale ocean current systems, providing a more physically consistent deep learning operator for the collaborative representation of multi-scale ocean dynamic processes. In terms of overall structure, KunPeng retains the design pattern of separating spatial and channel mixing, using a feedforward neural network as the hidden feature channel mixer, reducing computational load while maintaining good performance, enabling the model to be deployed on consumer-grade GPU devices and applied to a wider range of ocean prediction scenarios.

\subsection{Adaptability of Different Architectures to Ocean Environments}

In the comparative analysis of models, different architectures exhibit significant feature preferences in multi-element ocean prediction. In temperature prediction, GraphCast's advantage in the 10-200m shallow layer reveals the potential of graph neural networks in representing thermodynamic processes, as its hierarchical correlation mechanism may be more suitable for dynamic modeling of multi-scale heat exchange within the mixed layer. Meanwhile, the KunPeng model shows competitiveness in surface and deep-sea temperature prediction, suggesting the adaptability of its feature extraction module to the laws of fluid mechanics under a non-uniform hierarchical structure. Notably, KunPeng's absolute advantage in velocity field prediction (comprehensively leading in all 15-day prediction metrics) may stem from the dynamic feature pyramid constructed by deformable convolution, which captures the time-varying patterns of water movement through adaptive spatial receptive fields, making it more suitable for daily-scale dynamic process evolution than fixed graph structure modeling.

The divergence in salinity prediction metrics further reveals the differences in model characteristics: GraphCast's MAE advantage in the 30—150 m layer reflects its stable capture of salinity averages, while KunPeng's superior MSE and ACC metrics demonstrate its convergence advantage in prediction results. This difference may stem from the models' sensitivity discrepancy to outliers—the global information transfer of graph neural networks may amplify local disturbances, while KunPeng effectively suppresses prediction fluctuations through feature constraint mechanisms. Additional stratified metrics reinforce this finding (see Appendix~\ref{appendix:C}): Advantages of GraphCast across key metrics in sea surface height prediction is related to its surface feature fusion capability, while its limitations in mixed layer depth prediction expose the modeling bottleneck of spatial-channel coupling architectures for vertical stratification features. In contrast, KunPeng adopts a decoupled design, independently optimizing spatial topology and channel correlation, demonstrating stronger adaptability to vertical heterogeneity in the ocean.

\subsection{Further Exploration of Ocean Prediction Models}
\subsubsection{More Meteorological Coupling Data}

The ocean system serves as a critical interface for coupling Earth's surface spheres, and its dynamic and thermal processes are inherently governed by multi-sphere interactions. Atmospheric circulation drives surface circulation evolution through momentum and heat exchange at the air-sea interface, solar radiation flux dominates the thermal structure of the ocean mixed layer, and terrestrial freshwater input (such as runoff and glacier melting) significantly alters coastal salinity gradients and ocean layer stability. Future efforts should focus on developing dynamic assimilation techniques for multi-source meteorological-hydrological-radiation data, constructing a fully coupled land-sea-air-ice modeling framework to constrain multi-scale ocean processes with more comprehensive boundary conditions, enhancing prediction robustness in complex environments.  

\subsubsection{Long-Period Ocean Environmental Feature Capture}

The complexity of the ocean system is not only reflected in space complexity but also in its multi-time-scale coupled evolution characteristics. The current modeling framework based on a 10-year dataset can capture the influence of the physical field on interannual fluctuations (such as ENSO phase transitions), but significant limitations remain in representing decadal oscillations (such as the Pacific Decadal Oscillation, PDO). Future research should explore deep learning architectures that couple long and short cycles, separating signals of different time scales through time-frequency analysis modules and capturing the persistent influence of historical states with memory-enhanced networks. Simultaneously, embedding the intrinsic oscillation laws of the climate system into model prior knowledge through physical constraint mechanisms will enhance the generalization ability of slow-varying ocean processes under limited data conditions.

\subsubsection{Ensemble Forecasting}

The heterogeneity of current ocean numerical prediction models provides new opportunities for developing ensemble forecasting systems. The model revealed in this study indicate that different architectures have complementary representation capabilities for multi-scale ocean processes. An ensemble framework can dynamically allocate weights, emphasizing the heat exchange modeling advantages of graph neural networks in temperature-salinity prediction while prioritizing the dynamic feature extraction capabilities of deformable convolution in velocity field inversion. This multi-model collaborative mechanism not only achieves complementary fusion in feature space but also improves prediction robustness by avoiding systematic biases of single models, constructing a high-confidence, high-precision, low-computational-cost ensemble forecasting platform.

\section{Acknowledgement}

We thank Shanghai University for providing computing power trial that enabled us to complete the preliminary feasibility verification of the experiments. We thank the Pudong Branch of Shanghai Mobile for providing Ascend NPU servers for the subsequent experiments. We thank the Copernicus Marine Observation Center for providing ocean observation and reanalysis data free of charge.

\bibliographystyle{plainnat}  
\bibliography{references}

\newpage
\appendix

\section{Materials and Methods}
\label{appendix:A}
\subsection{Dataset}
\subsubsection{Data Scope}
This study focuses on the global ocean spatial domain with a temporal window spanning a continuous 12-year observation period from January 1, 2010 to December 31, 2021. We integrate three authoritative reanalysis datasets: the Global Ocean Ensemble Physics Reanalysis (GOEPR; \href{https://doi.org/10.48670/moi-00024}{https://doi.org/10.48670/moi-00024}), the Global Ocean Biogeochemistry Analysis and Forecast (GOBAF; \href{https://doi.org/10.48670/moi-00015}{https://doi.org/10.48670/moi-00015}), and the ERA5 atmospheric reanalysis dataset (\href{https://doi.org/10.24381/cds.adbb2d47}{https://doi.org/10.24381/cds.adbb2d47}). The GOEPR dataset provides three-dimensional ocean physical fields, including temperature (\(T\)), salinity (\(S\)), zonal current velocity (\(U\)), meridional current velocity (\(V\)), mixed layer depth (\(MLD\)), and sea surface height (\(SSH\)). Among these parameters, \(T\), \(S\), \(U\), and \(V\) are represented as 38-layer vertical profile data, while \(MLD\) and \(SSH\) are provided as two-dimensional horizontal fields. The ERA5 dataset delivers 2 m air temperature (\(T2M\)) and 10 m wind speed components (\(U10M\), \(V10M\)). All dynamic variables maintain a spatial resolution of 0.25$^\circ$×0.25$^\circ$, with temporal resolutions of daily scale for ocean physical and atmospheric parameters, while topographic data remain static fields.

\subsubsection{Data Preprocessing}
For the 38-layer provided by GOEPR, spanning depths of 0–420 m with vertical resolutions ranging from 0.5 m to 411.8 m, linear interpolation was applied to vertically resample the original non-uniform layers into 10 standardized isobaths: 1, 5, 10, 20, 50, 100, 150, 200, 300, and 400 m. Meanwhile the missing data in the Antarctic continental region were filled using spatial interpolation. The 30-layer ocean mask data from GOBAF were linearly interpolated to align with the same vertical layers and subsequently converted into binary categorical variables (0/1) via a ceiling function. Post-processing, the three-dimensional physical fields (temperature/salinity/zonal velocity/meridional velocity, \(T\)/\(S\)/\(U\)/\(V\)) formed a tensor of dimensions (10,720,1440), while two-dimensional planar data (mixed-layer depth/sea surface height, \(MLD\)/\(SSH\)) was structured as (1,720,1440). Topographic and atmospheric variables retained their original (1,720,1440) dimensionality.

To eliminate scale discrepancies among multi-source datasets during model training, dynamic variables were normalized using the z-score standardization method~\cite{RN26} (Equation~\ref{eqn:normalization}):

\begin{equation}
	X'_{tvchw} = \frac{X_{tvchw} - \mu_{vchw}}{\sqrt{\sigma^2_{vchw}}}
	\label{eqn:normalization}
\end{equation}

\noindent where $t \in T$ denotes the temporal dimension, $v$ represents variable type, $c$ corresponds to vertical layers, and $h$ and $w$ index latitude and longitude, respectively. $\mu_{vchw}$ and $\sigma^{2}_{vchw}$ are the spatiotemporal historical mean and variance for each variable-layer-location attribute, and $X'_{tvchw}$ is the dimensionless standardized value.

Zero-value features in oceanic physical fields lack statistical distinctiveness. Standardization transforms raw samples into a normalized distribution with zero mean and unit variance, thereby eliminating their original data characteristics. To ensure the statistical significance of zero-filled landmark regions (The probability of values lying beyond the $\mu \pm 3\sigma$ interval is less than 0.3\%, which indicates that the proportion of $0$ values within the masked region exceeds 99.7\% of the total $0$ values in the sample.), we implemented a post-standardization zero-padding strategy for terrestrial regions. This methodology effectively mitigates land signal interference in marine physical fields while maintaining mathematical consistency within the standardized feature space. Finally, through channel-wise fusion of multilayer variables and single-layer parameters, we constructed a spatiotemporal tensor of dimensions (46, 720, 1440), encompassing global coverage at 0.25$^\circ$ spatial resolution and daily temporal resolution.

\subsection{Model Methodology}
\subsubsection{Main Structure of the Model}

As illustrated in Figure~\ref{fig:pig1}(a), KunPeng is a deep learning architecture based on deformable convolution~\cite{RN27}, comprising three core modules: the Patch Embedding Module, the Feature Encoding-Decoding Module, and the Patch Restoration Module. These modules enable end-to-end feature learning through hierarchical cascading structure. The Patch Embedding Module employs $4 \times 4$ convolutional kernels to perform spatial downsampling on input tensors $(C, H, W)$, generating high-dimensional feature representations with dimensions $(C_1, H/4, W/4)$. Following an empirical formula $C_1 = \alpha C \ (\alpha \geq 4 \times 4)$, this module constructs an overcomplete representation space while preserving original information entropy, effectively reducing computational complexity of spatial mixing to $1/16$ of the original space. The Feature Encoding-Decoding Module adopts a symmetric encoding-decoding architecture and consists of an encoder and a decoder. The encoder contains two cascaded deformable convolution units: the first maintains $C_1$ channel dimensions, while the second compresses features to $(C_1 \times 2, H/8, W/8)$ through stride 2 convolution. Each unit integrates four deformable convolution operators whose deformation offsets dynamically adjust receptive fields via learnable parameters. The decoder adopts an inverse architecture, utilizing transposed convolutions to restore feature dimensions to $(C_1, H/4, W/4)$. Residual connections between the original embeddings and decoder outputs enable effective gradient flow, mitigating vanishing gradient issues. The Patch Restoration Module employs $4 \times 4$ transposed convolutions for spatial resolution recovery, mapping high-level features $(C_1, H/4, W/4)$ to target space $(C', H, W)$. Through channel pruning strategies (typically $C' = C - \Delta C, \ \Delta C \geq 1$), non-predictive dimensions such as seabed topographic topology layers are excluded, thereby streamlining the final representation space. 

\subsubsection{Longitude-Cyclic Deformable Convolution}

Global operators~\cite{RN28, RN29} establish long-range dependencies through homogeneous computational resource allocation. While demonstrating superior performance in meteorological modeling, these approaches struggle to adapt to the multiscale physical characteristics of marine environments. Oceanic processes simultaneously encompass: quasi-steady large-scale gyres (e.g., subtropical gyres with spatial scales $>1000$ km); mesoscale eddies (diameter $10$—$400$ km); and submesoscale structures like fronts/thermoclines ($<10$ km). The global attention mechanism applies equivalent computational weights to features across all scales, making it challenging for models with limited computational budgets to sufficiently learn fine-grained local features.

Deformable convolution enhances feature extraction by learning spatially-variant offsets that enable adaptive spatial adjustment of convolutional kernel sampling points based on input features. Compared to standard convolution with fixed grid sampling, this mechanism demonstrates superior capability in capturing localized abrupt features like oceanic fronts. However, traditional deformable convolution remains constrained by conventional boundary assumptions from image processing - its sliding kernels inherently treat image boundaries as independent entities through zero-padding or mirror-padding strategies for out-of-bound sampling. This fundamentally conflicts with the inherent longitudinal periodicity of marine environments (where the $-180^\circ$ and $180^\circ$ meridians are physically continuous), causing artificial discontinuities in cross-date-line current features and potentially generating spurious eddy closure signals.

To resolve this, we propose the Longitude-Cyclic Deformable Convolution Neural (LC-DCN, see Figure~\ref{fig:pig1}(c)). By reconstructing the spatial mapping mechanism of feature maps, we extend the convolutional kernel's sampling range into a ring-shaped topological structure along the east-west axis: when offsets guide sampling points across longitudinal boundaries, they automatically remap to corresponding positions on the antipodal edge. This innovation enables continuous modeling of hydrographic features between the western Pacific and eastern Atlantic.

\subsubsection{Loss Function Enhancement}

Meteorological large models commonly employ latitude-weighted loss functions~\cite{RN24, RN23, RN22} to compensate for grid distortion effects in high-latitude regions under Mercator projection. However, marine environment modeling faces two unique challenges: Firstly,significant spatial discontinuities caused by continental shelves, islands, and bathymetry result in NaN values for non-ocean areas, creating steep gradients at land-sea boundaries; Secondly, zero-padding strategies induce an imbalanced gradient descent rate, where model training progresses substantially faster in simple land-filled regions than in complex ocean areas, leading to spurious convergence phenomena. To address these issues, we propose a Masked Latitude-weighted L1 Loss (MLL1) function, whose mathematical formulation is shown in Equation~\ref{eq:loss}.

\begin{equation}
	L1_{tv} = \frac{1}{N_{\text{deep}} \times N_{\text{lat}} \times N_{\text{lon}}} \sum_{c=1}^{N_{\text{deep}}} \sum_{h=1}^{N_{\text{lat}}} \sum_{w=1}^{N_{\text{lon}}} M_{\text{ch}w} L_h \left| \hat{Y}_{\text{ch}w}^{tv} - Y_{\text{ch}w}^{tv} \right|
	\label{eq:loss}
\end{equation}

\noindent where \(\hat{Y}_{\text{ch}w}^{tv}\) and \(Y_{\text{ch}w}^{tv}\) denote the predicted value and ground truth value at the \(c\)-th channel, \(h\)-th latitude index, and \(w\)-th longitude index, respectively. \(M_{\text{ch}w}\) is a binary mask applied to mask out the non-ocean areas, and \(L_h\) is the latitude weight, which accounts for the distortion effect in high-latitude regions. \(L_h\) is defined as:

\begin{equation}
	L_h = N_{\text{lat}} \times \frac{\cos \phi_h}{\sum_{i=1}^{N_{\text{lat}}} \cos \phi_i}
	\label{eq:latitude_weight}
\end{equation}

\noindent where $\phi_h$ is the latitude angle. This coefficient is normalized by $\cos \phi$ to compensate for the area distortion of the Mercator projection. $M_{chw} \in \{0, 1\}$ is a multi-layer binary mask for marine and non-marine zones.

\subsubsection{Deformable Convolution-enhanced Multi-step Prediction}

Traditional Recurrent Neural Networks (RNN) achieve temporal modeling through hidden state transmission, but their sequential dependency leads to non-parallelizable computational graphs and linear memory complexity growth with sequence length. Although existing studies attempt to integrate RNN with Transformer architectures~\cite{RN30}, they remain constrained by computational resources and face practical deployment challenges in 0.25$^\circ$ high-resolution scenarios. 

DeepSpeed V3~\cite{RN31} proposes an innovative Multi-Token Prediction approach that innovatively constructs temporal regularization constraints through auxiliary prediction heads, introducing multi-step prediction objectives during training to enhance temporal reasoning capabilities while eliminating auxiliary modules during inference for zero parameter growth. However, its global attention mechanism demonstrates inadequate adaptability to marine multi-scale features.

This study proposes a Deformable Convolution-enhanced Multi-step Prediction (DC-MTP) method that inherits the training-inference structural decoupling advantage while improving prediction efficiency through local dynamic perception, as illustrated in Figure~\ref{fig:pig1}(d). The DC-MTP framework consists of a primary model and an auxiliary predictor. The primary model, built upon deformable convolution, takes samples from time step t as input and generates predictions for step $t + 1$. The auxiliary predictor comprises $k$ cascaded lightweight prediction units, each sharing the block embedding and block recovery modules of the primary model. Each unit sequentially processes the hidden layer outputs from the preceding module and inputs from time step $t + i$ to produce hidden layer features and outputs for step $t + i + 1$. Critically, the auxiliary predictor is activated solely during training to enhance temporal forecasting capabilities and is discarded during inference to ensure computational efficiency.

\subsection{Experimental Implementation}
\subsubsection{Equipment and Training}

The experiments were conducted using the Huawei Ascend 910b NPU cluster for model training. The hardware configuration consisted of an 8-card parallel architecture, with each card equipped with 64 GiB of memory. Parallel computing was implemented through the NPU adaptation interface \texttt{torch\_npu} in the PyTorch framework, utilizing the CANN 8.0.RC2 operator library. Due to the limited compatibility of the current operator library with certain native PyTorch interfaces, such as \texttt{torch.nn.functional.grid\_sample} and Fourier transform operations, a CPU co-processing strategy was adopted for the relevant operators in both the KunPeng model and the comparative models.

The dataset was divided into two independent subsets based on the time series: 3,652 samples from 2010 to 2019 were used as the training set, and 365 samples from 2021 were used as the test set. The total data size of the training set was 591 GiB, and the full training process took 134.73 hours to complete.

\subsubsection{Evaluation Metrics}

Existing studies commonly adopt latitude-weighted mean squared error (Latitude-weighted MSE) and latitude-weighted anomaly correlation coefficient (Latitude-weighted ACC) as evaluation metrics for meteorological prediction. In this study, to address the spatial heterogeneity characteristics of ocean data, we have improved the evaluation system by introducing a masked latitude-weighted approach. The masked latitude-weighted MSE is defined in Equation~\eqref{eq:masked_mse}, the latitude weight calculation method is given in Equation~\eqref{eq:latitude_weight}, and the masked latitude-weighted anomaly correlation coefficient is presented in Equation~\eqref{eq:acc}.

\begin{equation}
	MSE_{vt} = \frac{1}{N_{\text{deep}} \times N_{\text{lat}} \times N_{\text{lon}}} \sum_{c=1}^{N_{\text{deep}}} \sum_{h=1}^{N_{\text{lat}}} \sum_{w=1}^{N_{\text{lon}}} M_{\text{ch}w} L_h \left( \hat{Y}_{\text{ch}w}^{tv} - Y_{\text{ch}w}^{tv} \right)^2
	\label{eq:masked_mse}
\end{equation}

\begin{equation}
	ACC_{vt} = \frac{\sum_{c,h,w} M_{\text{ch}w} L_h \tilde{X}_{\text{vtchw}}^{\text{true}} \tilde{X}_{\text{vtchw}}^{\text{pred}}}{\sqrt{\sum_{c,h,w} M_{\text{ch}w} L_h \left( \tilde{X}_{\text{vtchw}}^{\text{true}} \right)^2} \sqrt{\sum_{c,h,w} M_{\text{ch}w} L_h \left( \tilde{X}_{\text{vtchw}}^{\text{pred}} \right)^2}}
	\label{eq:acc}
\end{equation}

where \(\tilde{X}_{\text{vtchw}}^{\text{true}} = X_{\text{vtchw}}^{\text{true}} - \overline{X}_{\text{cl,chw}}\) is the true anomaly value, and \(\tilde{X}_{\text{vtchw}}^{\text{pred}} = X_{\text{vtchw}}^{\text{pred}} - \overline{X}_{\text{cl,chw}}\) is the predicted anomaly value. \(\overline{X}_{\text{cl,chw}}\) represents the daily-mean of the physical-field samples from 2010 to 2019.

\subsubsection{Comparative Models}

To explore the adaptability of different algorithmic architectures in ocean prediction tasks, this study selects four representative Ocean-Atmosphere models for comparative analysis. The Pangu-Weather model based on Swin Transformer achieves global feature fusion through windowed attention mechanisms, with spatial complexity linearly related to input resolution, but requires adjustments for marine environments. Fourier neural operator models, FourCastNet and AI-GOMS, optimize spatial mixing processes using two-dimensional and one-dimensional Fourier transforms, respectively. While FourCastNet needs adaptation for ocean prediction, AI-GOMS, specifically designed for ocean temperature and salinity fields, only requires loss function adjustments. The graph neural network architecture, GraphCast, integrates coarse-to-fine-grained features through a multi-scale mesh graph message-passing network, but its atmospheric domain association graph needs reconfiguration for ocean element relationships. The KunPeng model in this study innovatively employs longitude-cyclic deformable convolutions to construct a feature pyramid structure, capturing spatial characteristics of ocean parameters through adaptive receptive field expansion. Notably, except for AI-GOMS, all comparative models originate from atmospheric prediction domains. This study ensures fairness in comparative experiments by reproducing their official implementations (see Appendix~\ref{appendix:D}) on a unified hardware platform and training dataset, adjusting input-output dimensions and physical constraints for ocean tasks.

\subsubsection{Hyperparameters}

The model is trained using the AdamW optimizer with an initial learning rate of \(1.25 \times 10^{-4}\), L2 weight decay coefficient of \(1 \times 10^{-4}\), momentum parameters \(\beta_1 = 0.9\) and \(\beta_2 = 0.999\), and numerical stability constant \(\epsilon = 1 \times 10^{-8}\). The hidden layer dimension in the network architecture is expanded from 768 to 1536 through linear projection, and the feedforward neural network adopts a two-layer perceptron structure with a hidden layer dimension four times the input layer. A batch size of 8 is set during training, and distributed training is performed on 8 NPUs using a data parallelism strategy, iterating for 20 epochs. The longitude-cyclic deformable convolution, based on deformable convolution v4~\cite{RN32}, uses a multi-head attention-like parallel computing architecture with 8 independent offset prediction heads, configured with a \(3 \times 3\) kernel, stride $1$, and padding $1$ to maintain feature map resolution stability. The DC-MTP integrates 5 auxiliary prediction branches. Normalization layers uniformly use LayerNorm, and the activation function employs the GELU nonlinear unit. The learning rate scheduler adopts the WarmupCosineLR strategy, with a total of 8000 iterations, including a 32-step warmup phase, and the minimum learning rate decays to \(1 \times 10^{-4}\). The input tensor dimensions are \((8, 6, 46, 720, 1440)\), corresponding to batch size, time steps, variable channels, latitude resolution, and longitude resolution, while the output tensor is adjusted to \((8, 6, 45, 720, 1440)\), where the time dimension 6 includes the main model prediction steps and 5 auxiliary prediction results. Auxiliary predictions are removed during inference, and the input tensor is \((1, 1, 46, 720, 1440)\).

\subsubsection{Optimization Details}

To enhance the training efficiency of large-scale ocean prediction models, this study adopts a three-tier optimization scheme: 
\textbf{1) ZeRO-2 Distributed Optimization}: Efficient memory utilization is achieved through DeepSpeed framework’s Zero Redundancy Optimizer (Stage 2)~\cite{RN33}, which partitions optimizer states and gradient tensors across 8 NPUs, reducing single-device memory usage to approximately 60\% of traditional data parallelism. 
\textbf{2) Mixed Precision Training}: A dual-precision system is constructed with FP32 master weights and FP16 computation copies. Forward propagation and gradient calculations use FP16, while critical operations such as layer normalization and loss functions retain FP32 precision. 
\textbf{3) Shared Memory Data Caching}: Leveraging DeepSpeed’s multi-process data parallelism, a shared memory cache pool with date granularity is designed. Training samples are mapped to 4000 memory blocks based on date indices, using an LRU (Least Recently Used) replacement strategy to enable cross-process data sharing, saving 20--40\% of training time depending on the model.

\subsubsection{Training Strategy}

Contemporary meteorological forecasting models often employ multi-phase training strategies to mitigate error accumulation. For instance, the Pangu-Weather model~\cite{RN24} employs a four-stage cascaded prediction framework: training multiple submodels with distinct input-output time intervals, decomposing long-term predictions into iterative steps via a reverse prediction chain to reduce cumulative errors. The Fuxi v2 model~\cite{RN34} proposes a frequency-splitting collaborative strategy, where its primary model predicts low-frequency components at 5-frame intervals, while an auxiliary interpolation model reconstructs high-frequency details, achieving continuous prediction through frequency-domain fusion. Although such strategies significantly enhance accuracy, they introduce architecture-dependent interference. To establish a standardized benchmarking framework, this study strictly controls experimental variables by avoiding all multi-model collaboration or iterative refinement strategies. Through raw end-to-end single-model prediction, we ensure performance comparisons across algorithms in marine scenarios solely reflect core architectural differences, while constraining training resource consumption within 1,500 GPU-hours of single-card equivalent compute budget. This design adheres to the controlled variable principle, providing unbiased evaluation criteria for operator selection in ocean prediction models.

\FloatBarrier

\section{Optimization Details}
\label{appendix:B}

In the research on improving the training efficiency of large-scale ocean prediction models, we propose a three-stage linkage optimization scheme, which integrates three key technologies: ZeRO-2 distributed optimization, mixed-precision training, and shared memory data caching. The following systematically elaborates on each optimization scheme.

\subsection{ZeRO-2 Distributed Optimization}

Based on the Zero Redundancy Optimizer (Stage 2) of the DeepSpeed framework, ZeRO-2 significantly reduces the memory occupation of a single device by means of a partitioned storage mechanism for gradients and optimizer states. In traditional data parallel training, each device needs to store the gradients and optimizer states (such as momentum and adaptive learning rate parameters) completely, resulting in low memory utilization. In this study, an 8-NPU device cluster is used. During the backpropagation stage, the gradient tensors are partitioned according to rules, and each device only stores the corresponding partitions; the optimizer states are also stored in a distributed manner. This strategy reduces the memory occupation of a single card, allows breaking through the limitations of hardware resources, supports the training of larger-scale models, and fully improves the collaborative efficiency of multiple devices through parallel computing. Experiments show that ZeRO-2 has good scalability, and an increase in the device scale can further reduce the memory pressure and accelerate the training.

\subsection{Mixed-Precision Training}

In this study, a dual-precision system of FP32 main weights and FP16 calculation replicas is constructed, taking into account both training speed and numerical stability. Although traditional FP32 training has reliable accuracy, its computational efficiency is relatively low; while FP16 can significantly increase the speed, it has the risks of gradient underflow and accuracy loss. In response to this, FP16 format is used for forward propagation and gradient calculation, and FP32 accuracy is retained for key operations (such as layer normalization and loss function calculation). It should be specifically noted that the Fourier transform in the adaptive Fourier neural operator needs to support complex number operations, so it is switched to the FP32 mode during forward propagation. This strategy not only uses FP16 to accelerate the main body of the calculation but also ensures the numerical accuracy of key links through local accuracy improvement. At the same time, the study further introduces the \texttt{offload\_optimizer} technology, which migrates the non-core calculation tasks of the optimizer (such as parameter updates and state maintenance) to the CPU for processing, releasing the memory pressure of the GPU and enabling it to focus on high-load tasks such as tensor operations. The heterogeneous collaboration between the CPU and the GPU effectively avoids training interruptions caused by memory bottlenecks.

\subsection{Shared Memory Data Caching}

In view of the data-intensive characteristics of ocean prediction models, this study, based on the DeepSpeed multi-process architecture, constructs a shared memory cache pool with the granularity of dates. The hard disk I/O operations have become a bottleneck in training efficiency due to frequent reading. The cache pool maps the training samples to 4,000 memory blocks according to the date index, reducing I/O operations and improving training efficiency. When loading data, the cache pool is accessed preferentially, and the memory data is directly read after a hit. The Least Recently Used (LRU) replacement strategy is adopted for cache management, with the algorithm logic presented in Algorithm~\ref{alg:lru_cache}.

\begin{algorithm}
	\caption{Least Recently Used (LRU) Cache}
	\label{alg:lru_cache}
	\begin{algorithmic}[1]
		\State \textbf{Input:} Cache capacity $C$, request sequence $R=(r_1, r_2, \dots, r_n)$
		\State \textbf{Output:} Hit/miss results $H=(h_1, \dots, h_n)$ with numerical values, and implicit final cache state
		\State $cache \gets empty\quad dictionary (key: item, value: counter)$
		\State $H \gets empty\quad list$ \Comment{Initialize hit/miss results list with numerical values}
		\State $current\_size \gets 0$ \Comment{Track the number of items currently in the cache}
		\For{$i \gets 1$ \textbf{to} $n$} \Comment{Process each request in the sequence}
		\State $request \gets r_i$ \Comment{Current request to be processed}
		\If{$request \in cache$}
		\State $hit\_counter \gets cache[request]$ \Comment{Get current counter before reset}
		\State $H.append(hit\_counter)$ \Comment{Record hit counter value}
		\State $cache[request] \gets 0$ \Comment{Reset hit item’s counter}
		\For{$each\quad item \in cache$}
		\If{$item = request$}
		\State $cache[item] \gets cache[item] + 1$
		\EndIf
		\EndFor
		\Else
		\State $H_{value} \gets 0$ \Comment{Default value for misses}
		\If{$current\_size < C$} \Comment{Case 2a: Cache not full}
		\State $cache[request] \gets 0$ \Comment{Add new item with counter 0}
		\State $current\_size \gets current\_size + 1$ \Comment{Update cache size}
		\State $H_{value} \gets 0$ \Comment{Record new item’s initial counter}
		\Else \Comment{Case 2b: Cache full, evict LRU item}
		\State $max\_counter \gets -1$
		\State $evict\_item \gets None$
		\For{each $(item, counter) \in cache.\text{items}()$}
		\If{$counter > max\_counter$}
		\State $max\_counter \gets counter$
		\State $evict\_item \gets item$
		\EndIf
		\EndFor
		\State $H_{value} \gets max\_counter$ \Comment{Record evicted item’s counter}
		\State $del cache[evict\_item]$ \Comment{Evict the item}
		\State $cache[request] \gets 0$ \Comment{Add new item}
		\EndIf
		\State $H.\text{append}(H_{value})$
		\For{each $item \in cache$}
		\If{$item = request$}vc
		\State $cache[item] \gets cache[item] + 1$
		\EndIf
		\EndFor
		\EndIf
		\EndFor
		\State \textbf{Return} $H$ \Comment{Return hit/miss results and implicit final cache state}
	\end{algorithmic}
\end{algorithm}

\FloatBarrier

\section{Comparative Models}
\label{appendix:D}

This section conducts a systematic analysis of the architectural design and implementation schemes of major meteorological and oceanographic prediction models. We have reproduced all comparative models following a technical roadmap of ``core module reconstruction + computational process optimization,'' with framework-specific adaptations: FourCastNet implements Fourier computation with CPU collaborative operations, and GraphCast achieves cross-framework migration from JAX to PyTorch. Based on the open-source status of the code, our implementations reference partial pseudocode from Pangu, core implementations from FourCastNet, and reproduce other models relying on descriptions from original papers. Training details are summarized in Table~\ref{tab:training_details}.
The code is shown in \href{https://github.com/kbdsbx/loem}{https://github.com/kbdsbx/kunpeng}.%

\begin{table}[h]
	\centering
	\caption{Implementation and Training Parameters of the Comparative Models}
	\label{tab:training_details}
	\begin{tabular}{lccccc}
		\toprule
		& Pangu & FourCastNet & AI-GOMS & GraphCast & KunPeng \\
		\midrule
		Core Algorithm & 3DEST & 2D-AFNO & 1D-AFNO & GNN & LC-DCN \\
		Initial Learning Rate & $1 \times 10^{-5}$ & $5 \times 10^{-4}$ & $5 \times 10^{-4}$ & $1 \times 10^{-3}$ & $1.25 \times 10^{-4}$ \\
		Regularization Term & $3 \times 10^{-6}$ & $0.001$ & $0.01$ & $0.1$ & $1 \times 10^{-4}$ \\
		Hidden Layer Size & 192 & 768 & 768 & 192 & 768 \\
		Training Epochs & 20 & 20 & 20 & 20 & 20 \\
		Training Time (h) & 28.4 & 23.8 & 84.83 & 85.27 & 134.73 \\
		Tranining MLL1 & 0.26 & 0.21 & 0.18 & 0.16 & 0.21 \\
		\bottomrule
	\end{tabular}
\end{table}

\subsection{Pangu-Weather Model}

The Pangu-Weather model\cite{RN24} adopts a 3D Earth-Specific Transformer (3DEST) architecture. Compared to traditional 2D Transformers, this model employs a stride strategy of $2\times4\times4$ to independently encode embeddings for 3D meteorological variables, followed by integration of the encoded blocks. To adapt to Earth's meteorological data characteristics, the research team enhanced the positional encoding mechanism of the Swin Transformer by replacing relative positional biases with Earth-specific positional biases. This design preserves the translation invariance of spatial mixing processes and significantly reduces the scale of bias parameters through a parameter-sharing mechanism while enhancing reusability. The model utilizes an encoder-decoder architecture, which effectively extracts high-resolution detailed features and achieves precise reconstruction. 

For training strategy, Pangu innovatively implements a four-stage cascaded prediction framework: four submodels with identical architectures are trained separately for input-output intervals of 1/3/6/24 time frames. During prediction, multi-step tasks are decomposed through an inverse prediction chain. For example, to predict 107 frames:
The 24-frame model predicts 4 times to reach 96 frames;
The 6-frame model advances to 102 frames;
The 3-frame model extends to 105 frames;
The 1-frame model completes the final 2 predictions.
This approach achieves the target in only 8 iterations, effectively minimizing error accumulation.

During model reproduction, we retained the core hyperparameter settings:
Initial learning rate: $1\times10^{-5}$;
L2 regularization coefficient: $3\times10^{-6}$;
Hidden layer dimension: 192;
Learning rate scheduler: WarmupCosineLR;
Loss function: Optimized to MLL1 loss.
To ensure fairness in comparative experiments, the original multi-stage cascaded framework was disabled during reproduction. After 28.4 hours of training, the model achieved a 58.1\% reduction in loss, with the final MLL1 loss stabilizing at 0.26.

\subsection{FourCastNet}

FourCastNet\cite{RN22} is constructed based on the Adaptive Fourier Neural Operator, and it realizes the interaction of global meteorological information through the transformation of frequency-domain features. Its core operations consist of two stages: feature space mixing and reconstruction. First, in the spatial mixing stage, the model maps the features of the hidden layer to the frequency-domain complex space through the two-dimensional Fast Fourier Transform (FFT). Subsequently, a diagonal Multi-Layer Perceptron (MLP) is employed to cross-mix the real and imaginary part features, and its mathematical expression is:

\begin{align}
	Y^{\text{$\Re$}}  &= g(X^{\text{$\Re$}} * W^{\text{$\Re$}} - X^{\text{$\Im$}} * W^{\text{$\Im$}} + b^{\text{$\Im$}}) \label{eq:y_real} \\
	Y^{\text{$\Im$}} &= g(X^{\text{$\Re$}} * W^{\text{$\Im$}} + X^{\text{$\Im$}} * W^{\text{$\Re$}} + b^{\text{$\Im$}}) \label{eq:y_imag}
\end{align}

Let \(X\) denote the input tensor, \(W\) the weight matrix, $b$ the bias term,  
$\Re$ and $\Im$ the real and imaginary components of complex-valued features, $g$ the activation function,  
and $\cdot$ the matrix multiplication operator. After feature mixing, the spectral features are reconstructed  
back to the parameter space via the inverse Fast Fourier Transform (iFFT).  
The model architecture employs an asymmetric encoder-decoder design:  
Encoder is composed of stacked multi-level Adaptive Fourier Neural Operators (2D-AFNO).
Decoder utilizes lightweight linear projection layers to achieve spatial dimension mapping.

During model reproduction, we strictly adhered to the official implementation. To address hardware constraints  
of the Ascend NPU (specifically, the CANN 8.0.RC2 operator library lacks support for \texttt{torch.fft} operator  
groups and imposes 16-bit complex number precision limits), we implemented a heterogeneous computing strategy:  
FFT/iFFT operations were explicitly offloaded to the CPU for execution.
32-bit complex number precision quantization was applied to ensure computational accuracy.
Core hyperparameters retained the original paper's settings:
Initial learning rate: $5\times10^{-4}$;
L2 regularization coefficient: $0.01$;
Hidden layer dimension: $768$.
After 23.8 hours of training, the model achieved a 67.9\% reduction in loss, with the final MLL1 loss stabilizing at $0.21$.

\subsection{AI-GOMS}

AI-GOMS\cite{RN20} (Large AI-Driven Global Ocean Modeling System) is a model specifically designed for oceanographic prediction. Compared to FourCastNet, its core innovation lies in employing the 1D Adaptive Fourier Neural Operator (1D-AFNO) to process oceanographic data and enhancing the model's ability to reconstruct discontinuous oceanographic features through a random masking strategy. During block embedding, the model executes a random block deletion strategy, retaining only a subset of blocks for learning, and reconstructs complete oceanographic features at the output stage.  
The AI-GOMS architecture adopts an asymmetric encoder-decoder design:  
Encoder is composed of stacked multi-level 1D-AFNO operators.
Decoder is a lightweight linear layer that can be replaced with task-specific decoders for diverse downstream applications, such as ocean chlorophyll prediction or wave height forecasting.

In our code reproduction, we referenced the original literature. Since the literature did not specify hyperparameters, we maintained consistency with FourCastNet:
Initial learning rate: $5\times10^{-4}$;
L2 regularization coefficient: $0.01$;
Hidden layer dimension: $768$.
After 84.83 hours of training, the model achieved a 70.6\% reduction in loss, with the final MLL1 loss stabilizing at $0.18$.

\subsection{GraphCast}

GraphCast\cite{RN23} adopts an architecture based on graph neural networks and achieves the efficient propagation of meteorological features by constructing a multi-scale geophysical topological graph. The model takes the icosahedron as the initial topological structure. Through six recursive subdivision processes, in each process, the triangular faces are divided into four equal parts. Finally, a multi-resolution topological graph containing 40,962 vertices, 81,920 faces, and 245,760 directed edges is generated. It is worth noting that by retaining the edge connections of all subdivision levels, the model constructs a composite topological structure with a total of 327,660 directed edges to enhance the multi-scale feature representation ability. 
The model architecture consists of three modules: an encoder, a processor, and a decoder.  
Encoder: Maps the meteorological tensor data to the feature space of the nodes in the topological graph, while embedding the spatial coordinate information of each node.
Processor: Conducts message passing through a 16-layer graph neural network and uses directed edges to achieve local feature interaction and global information fusion between nodes.
Decoder: Performs the inverse mapping process, reconstructing the graph-structured data into three-dimensional meteorological tensor data.

In terms of implementation, this study has reconstructed the core algorithm of the original Jax version based on the PyTorch framework, and removed the components coupled with the TypeGraph framework to improve code maintainability. During training, the initial learning rate is set to $1\times10^{-3}$, the L2 regularization coefficient is $0.1$, and the hidden layer dimension is $192$. After 85.27 hours of training, the model loss has decreased by 46.8\%, and the final MLL1 loss stabilizes at $0.16$.

\FloatBarrier

\section{Environmental Variables and Masks}
\label{appendix:E}

This section presents all the variables designed in the experiment, which involve three data sources, namely the \href{https://doi.org/10.48670/moi-00024}{Global Ocean Ensemble Physics Reanalysis (GOEPR)}, the \href{https://doi.org/10.48670/moi-00015}{Global Ocean Biogeochemistry Analysis and Forecast (GOBAF)}, and the \href{https://doi.org/10.24381/cds.adbb2d47}{ERA5 Atmospheric Reanalysis Dataset}.See Table~\ref{tab:variables} for details.

\begin{table}[H]
	\centering
	\caption{Experimental Variables}
	\label{tab:variables}
	\begin{tabular}{l l l l}
		\toprule
		\textbf{Variable} & \textbf{Layer} & \multicolumn{1}{l}{\textbf{Details}} & \textbf{Source} \\
		\midrule
		T       & 10  & Sea temperature ($^\circ$C)                      & GOEPR \\
		S       & 10  & Sea salinity (PSU)                        & GOEPR \\
		U       & 10  & Sea stream zonal velocity (ms$^{-1}$)   & GOEPR \\
		V       & 10  & Sea stream meridional velocity (ms$^{-1}$) & GOEPR \\
		MLD     & 1   & Mixed layer depth (m)                     & GOEPR \\
		SSH     & 1   & Sea surface height (m)                    & GOEPR \\
		T2M     & 1   & Air temperature at 2m ($^\circ$C)               & ERA5 \\
		U10M    & 1   & Wind meridional velocity 10m (ms$^{-1}$) & ERA5 \\
		V10M    & 1   & Wind zonal velocity 10m (ms$^{-1}$)   & ERA5 \\
		deptho  & 1   & Digital elevation and bathymetry (m)      & GOBAF \\
		mask    & 10  & Marine mask (0/1)                        & GOBAF \\
		\bottomrule
	\end{tabular}
\end{table}

\section{Experimental Comparison Results}
\label{appendix:C}

This section presents the comparison of the indicators for the 15-day prediction of all 45 layers of ocean/atmosphere environmental variables in the experiment. 

\begin{figure}[htbp]
	\centering
	\includegraphics[width=\linewidth]{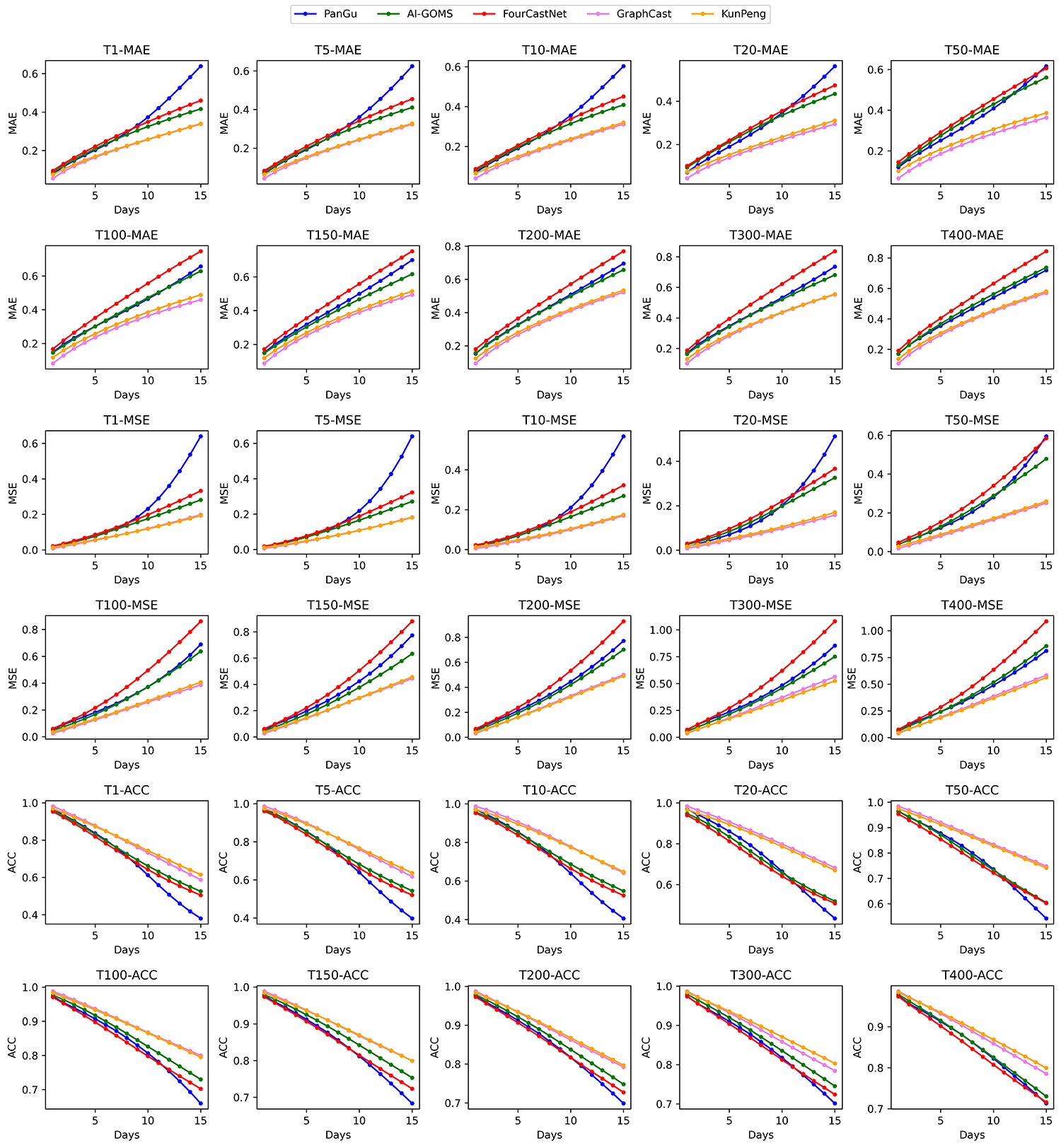}
	\caption{Comparison of 15 - day Temperature Predictions}
	\label{fig:15_day tempre}
\end{figure}

\begin{figure}[htbp]
	\centering
	\includegraphics[width=\linewidth]{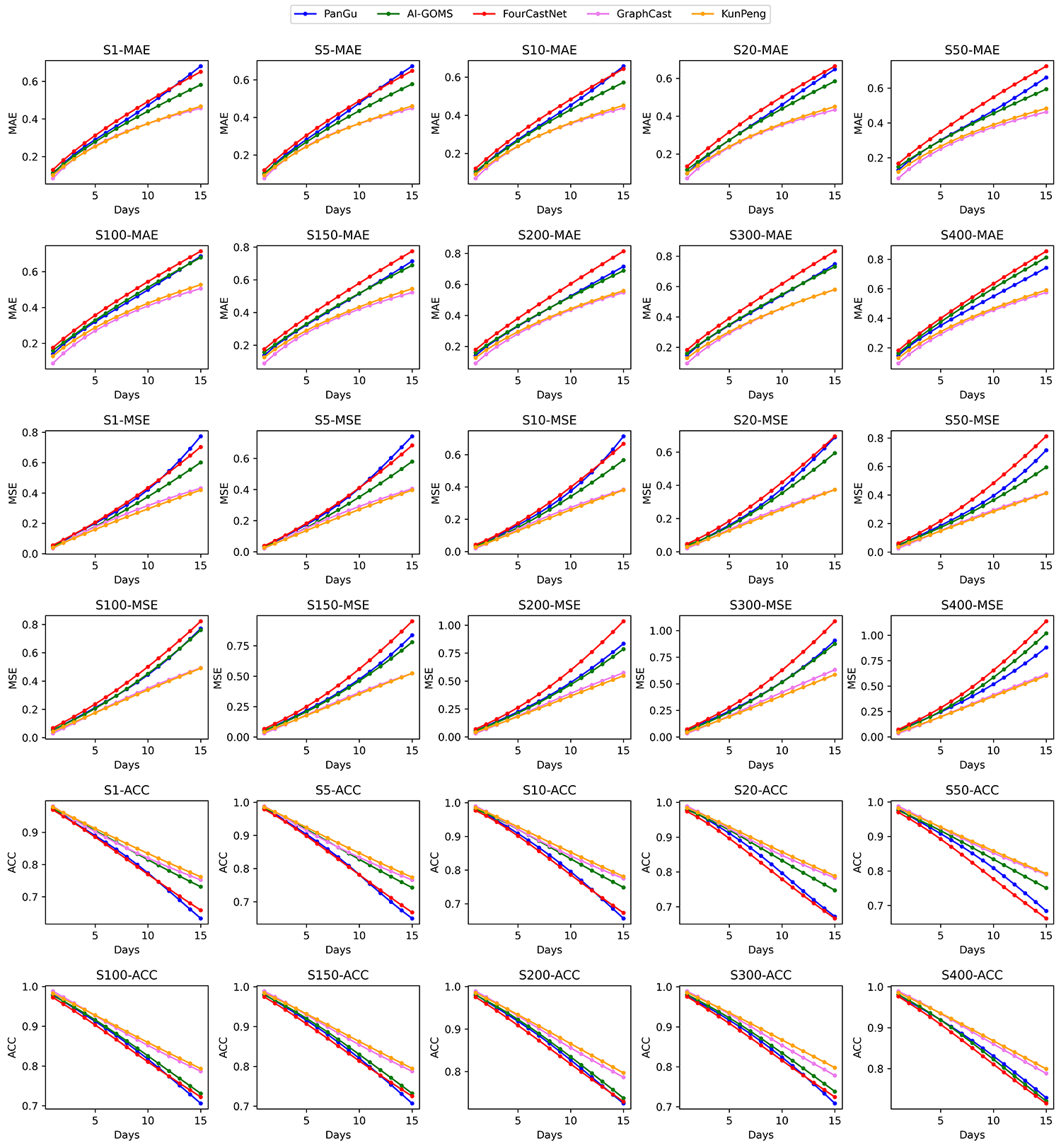}
	\caption{Comparison of 15 - day Salinity Predictions}
	\label{fig:15_day salpre}
\end{figure}

\begin{figure}[htbp]
	\centering
	\includegraphics[width=\linewidth]{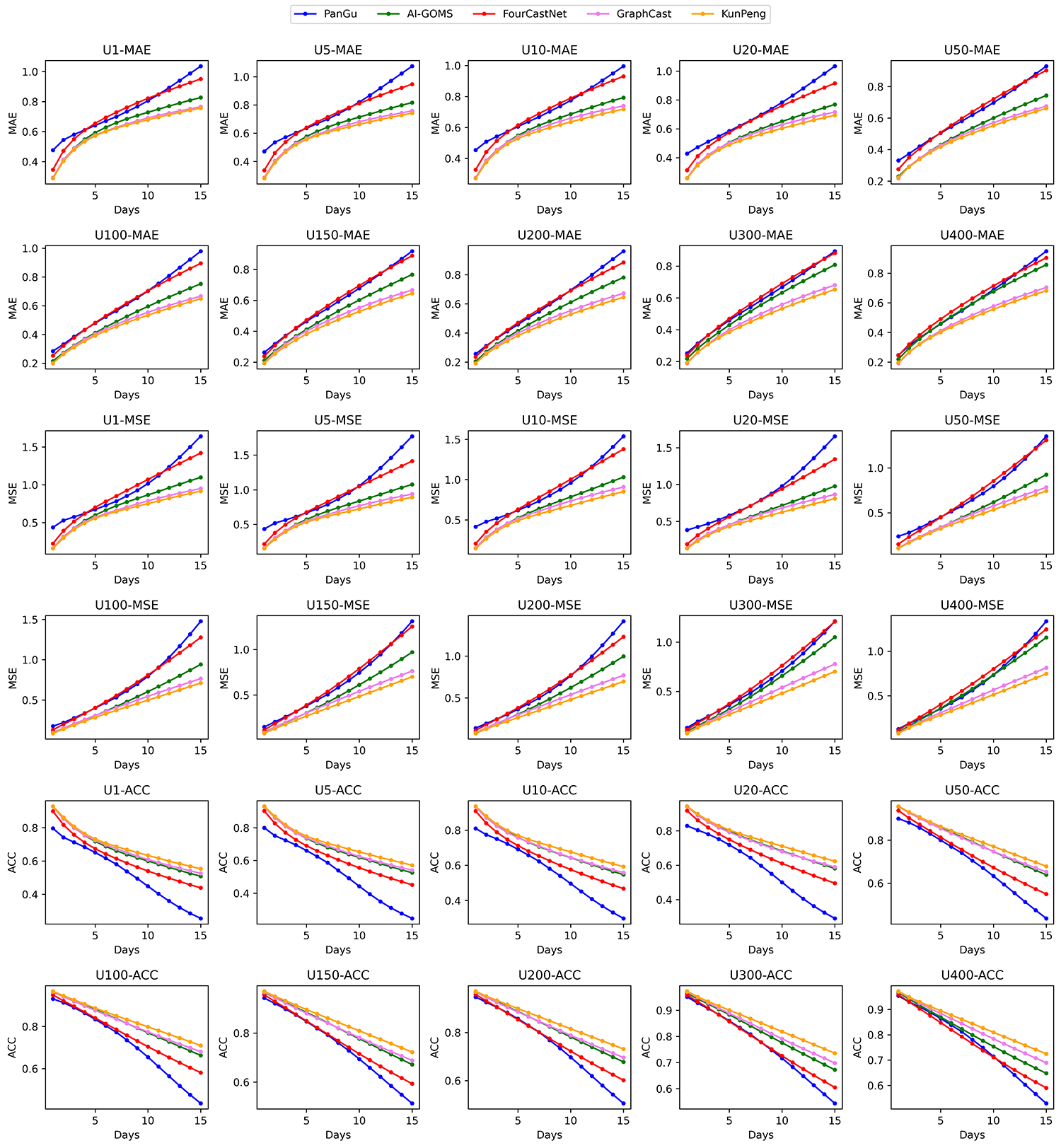}
	\caption{Comparison of 15 - day Meridional Velocity Predictions}
	\label{fig:15_day ravepre}
\end{figure}

\begin{figure}[htbp]
	\centering
	\includegraphics[width=\linewidth]{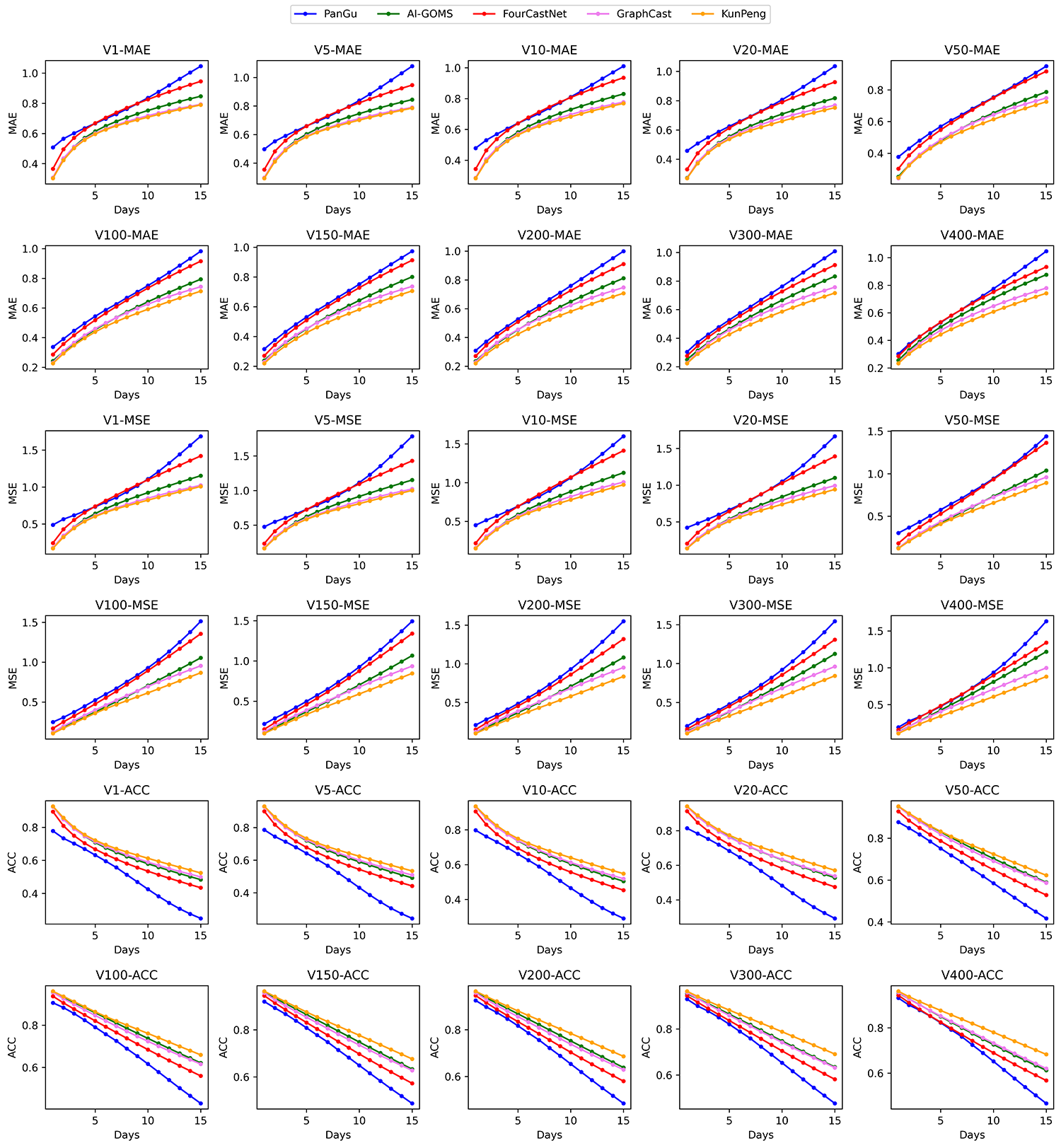}
	\caption{Comparison of 15 - day Zonal Velocity Predictions}
	\label{fig:15_day zonvelpre}
\end{figure}

\begin{figure}[htbp]
	\centering
	\includegraphics[width=0.8\linewidth]{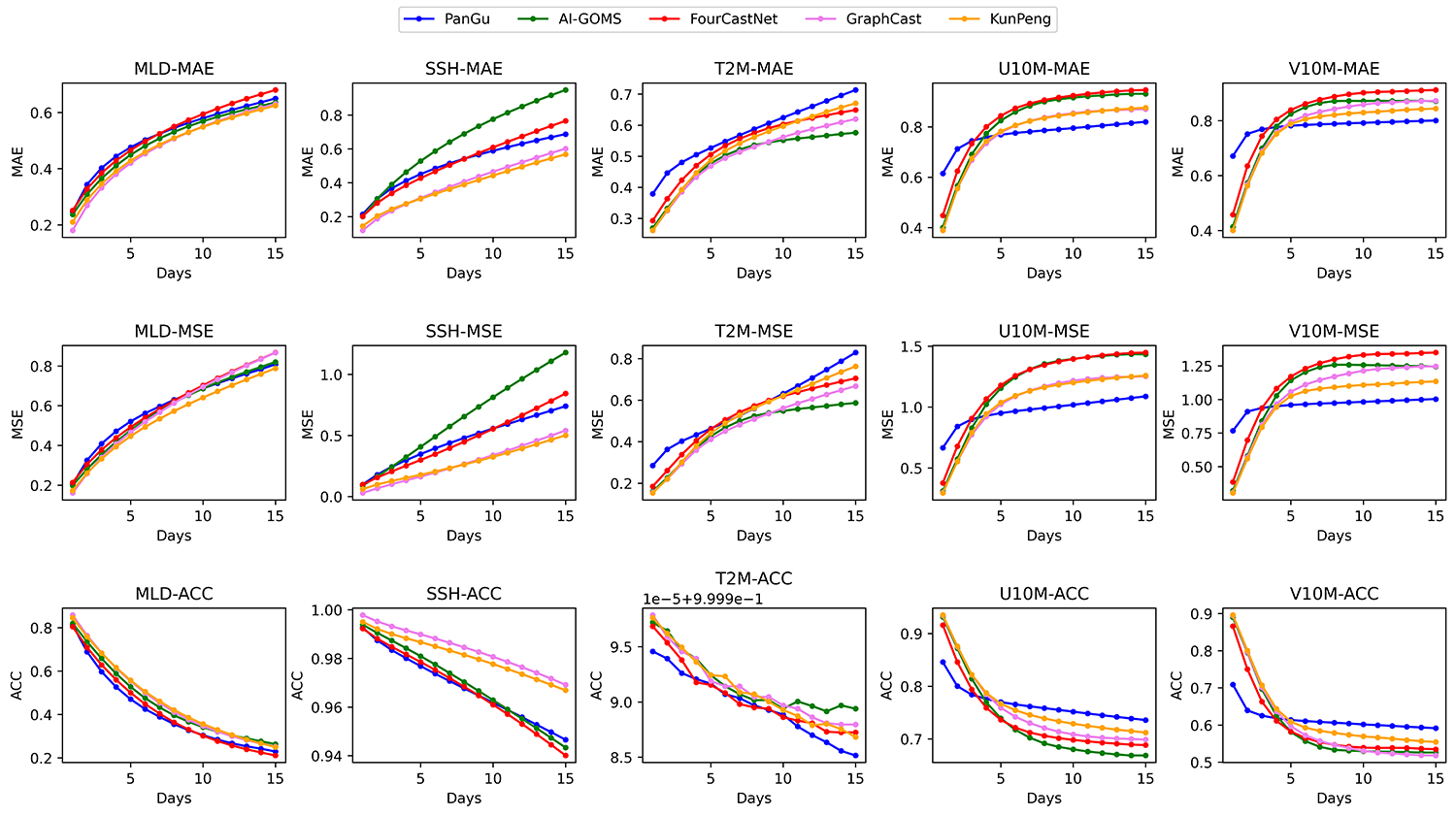}
	\caption{Comparison of Planar Indicators. \(MLD\) stands for Mixed Layer Depth, \(SSH\) for Sea Surface Height Anomaly, \(T2M\) for Air Temperature at 2 m above the Sea Surface, \(U10M\) for the East Wind Speed at 10 m above the Sea Surface, and \(V10M\) for the North Wind Speed at 10 m above the Sea Surface.}
	\label{fig:15_day pre}
\end{figure}

\FloatBarrier

\section{Classification Results}
\label{appendix:F}
\subsection{Error Accumulation Process}

\begin{figure}[htbp]
	\centering
	\includegraphics[width=\linewidth]{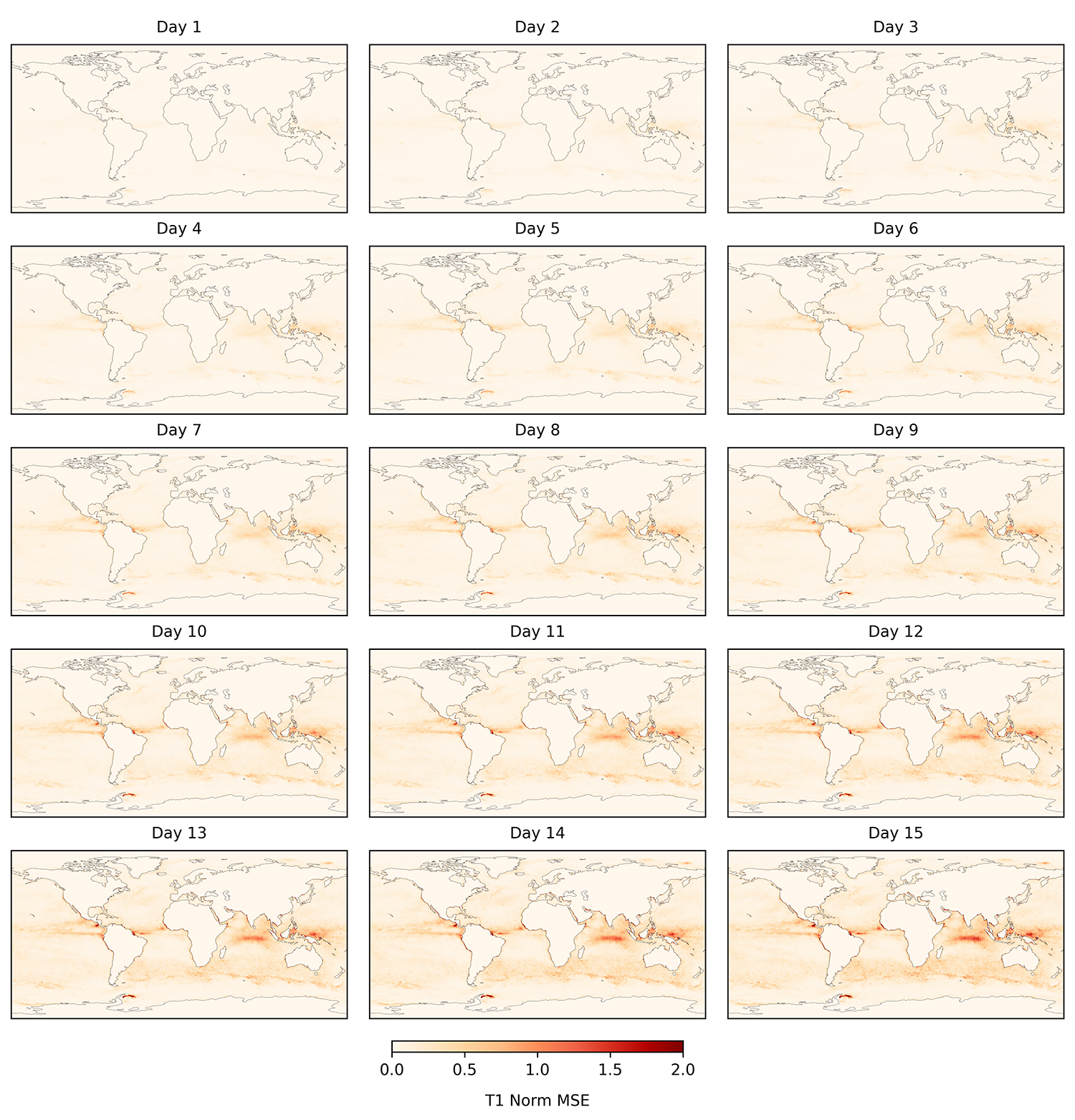}
	\caption{15-day normalized mean squared error (MSE) of sea surface temperature at 1 m depth in the KunPeng model, based on the 2021 test dataset.}
	\label{fig:E1}
\end{figure}

\begin{figure}[htbp]
	\centering
	\includegraphics[width=\linewidth]{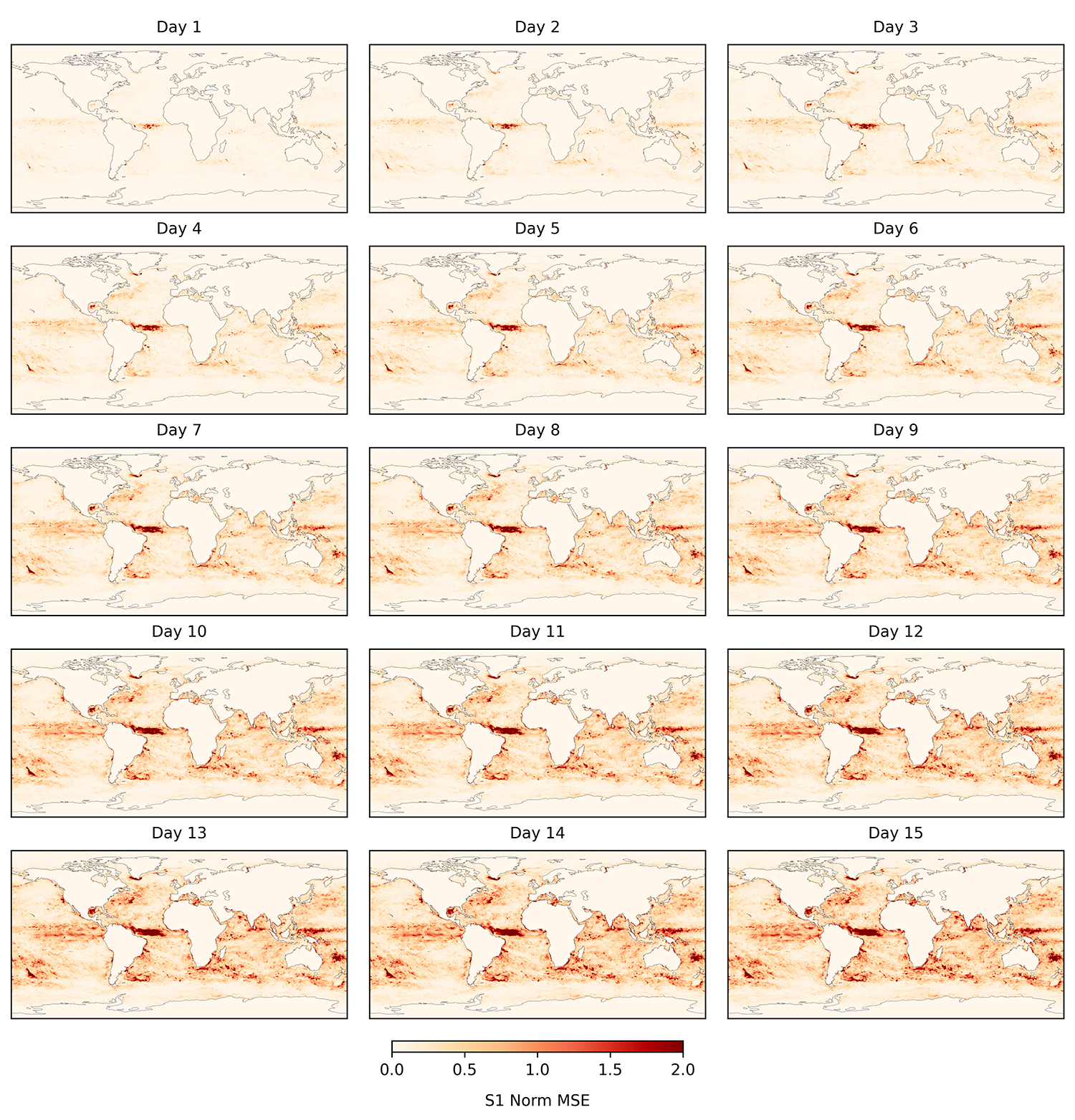}
	\caption{15-day normalized mean squared error (MSE) of sea surface salinity at 1 m depth in the KunPeng model, based on the 2021 test dataset.}
	\label{fig:E2}
\end{figure}

\begin{figure}[htbp]
	\centering
	\includegraphics[width=\linewidth]{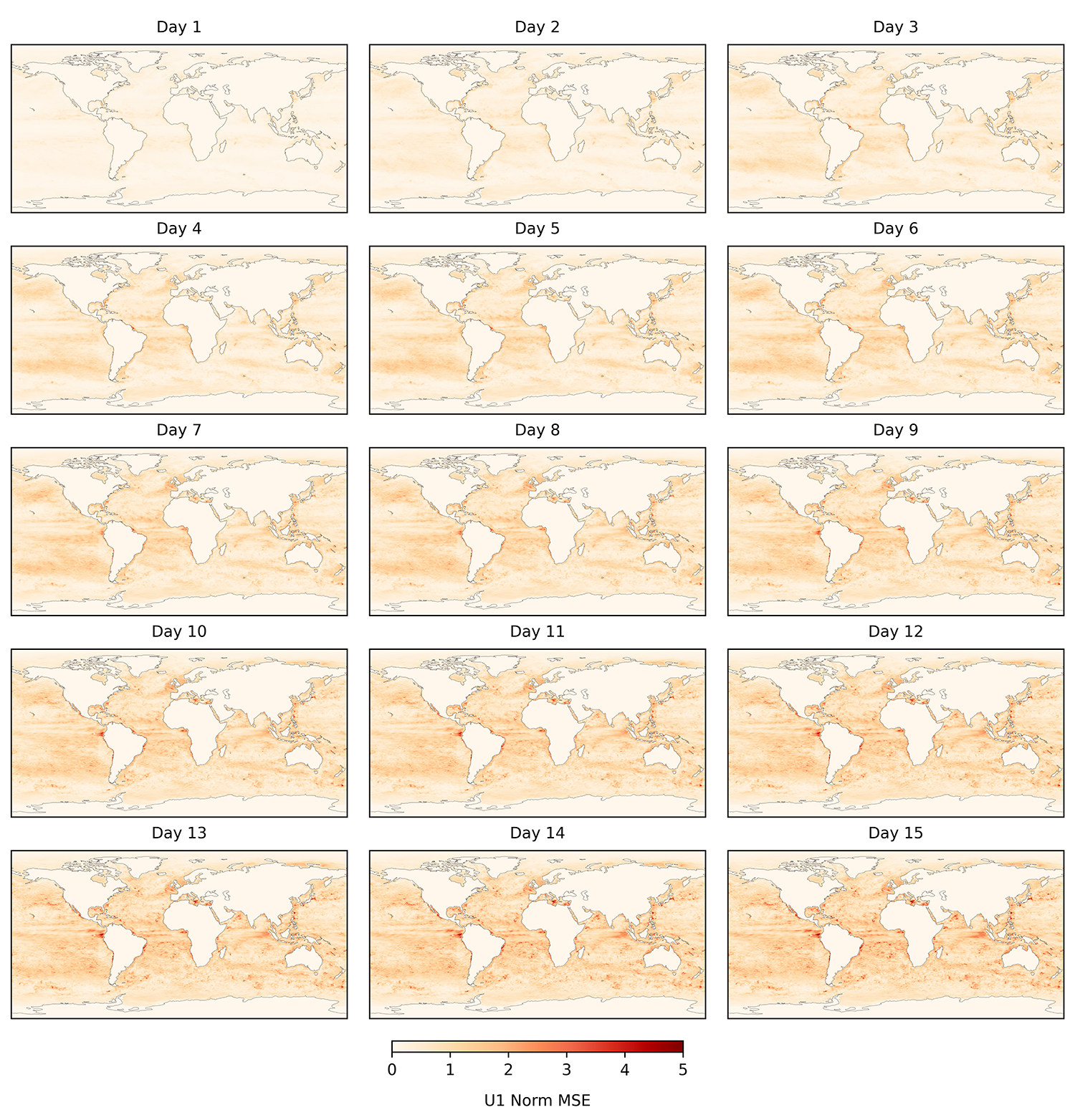}
	\caption{15-day normalized mean squared error (MSE) of meridional current velocity at 1 m depth in the KunPeng model, based on the 2021 test dataset.}
	\label{fig:E3}
\end{figure}

\begin{figure}[htbp]
	\centering
	\includegraphics[width=\linewidth]{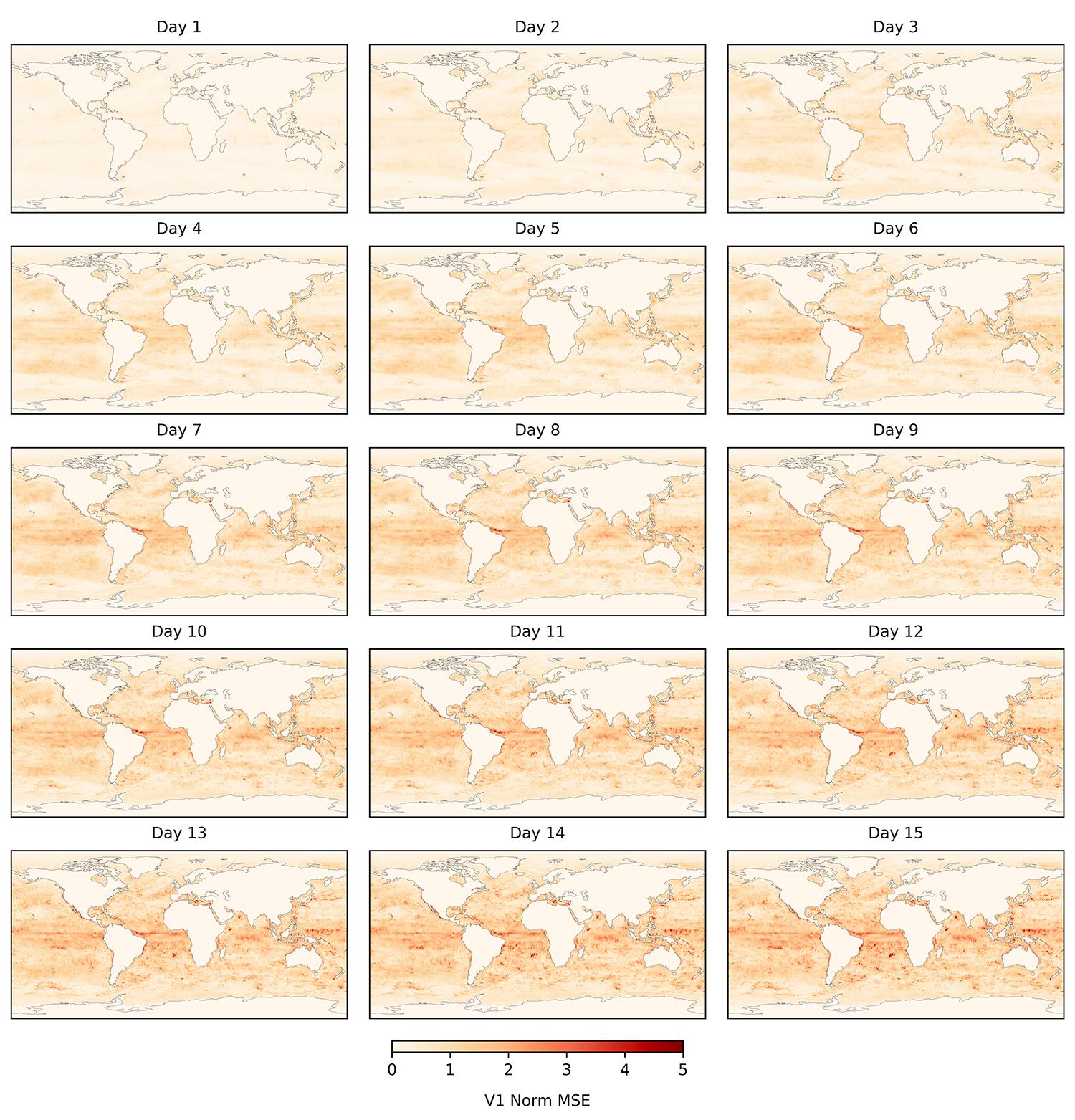}
	\caption{15-day normalized mean squared error (MSE) of zonal current velocity at 1 m depth in the KunPeng model, based on the 2021 test dataset.}
	\label{fig:E4}
\end{figure}

Figures~\ref{fig:E1}, \ref{fig:E2}, \ref{fig:E3}, \ref{fig:E4} illustrate the error accumulation process of the 15-day mean sea surface MSE for the KunPeng model. The statistics are based on the 15-day error averages from \texttt{2021-01-01T00:00:00Z UTC} to \texttt{2021-12-16T00:00:00Z UTC}.

From the perspective of marine environmental variables, the error accumulation rate for the current velocity field is notably higher than that for the temperature and salinity fields. Regionally, the spatial patterns of error accumulation vary across different environmental variables. For the velocity field, errors accumulate uniformly across global scales. In contrast, temperature-related errors are predominantly concentrated in the following regions: the Amazon River estuary, the equatorial Atlantic, the western equatorial Pacific near the Americas, the central Indian Ocean, the eastern equatorial Pacific near Southeast Asia, the Weddell Sea sector of Antarctica, and coastal zones adjacent to continental landmasses. Possible contributing factors to the pronounced temperature prediction errors in these regions may include: intense solar radiation in equatorial zones, large-scale ocean circulation dynamics, altered heat capacity due to freshwater influx from river discharges, and land-sea thermal coupling effects. 

For salinity variables, significant error accumulation is observed in the equatorial Atlantic, the southern Greenland Sea, the Gulf of Mexico, the eastern waters of New Zealand, and the eastern equatorial Pacific near Southeast Asia. Key drivers for these salinity prediction inaccuracies likely involve freshwater dilution from riverine inputs, glacial meltwater intrusions, and evaporation anomalies induced by solar radiation.

To enhance the prediction accuracy of temperature and salinity fields, it is recommended to incorporate spatial and atmospheric indicators (e.g., solar radiation, atmospheric humidity, wind speed, and precipitation) and refine the coupling processes between oceanic and continental characteristics. Such optimizations could mitigate systematic biases and improve model performance in critical marine regions.

\FloatBarrier

\subsection{Mean Bias}

\begin{figure}[htbp]
	\centering
	\includegraphics[width=\linewidth]{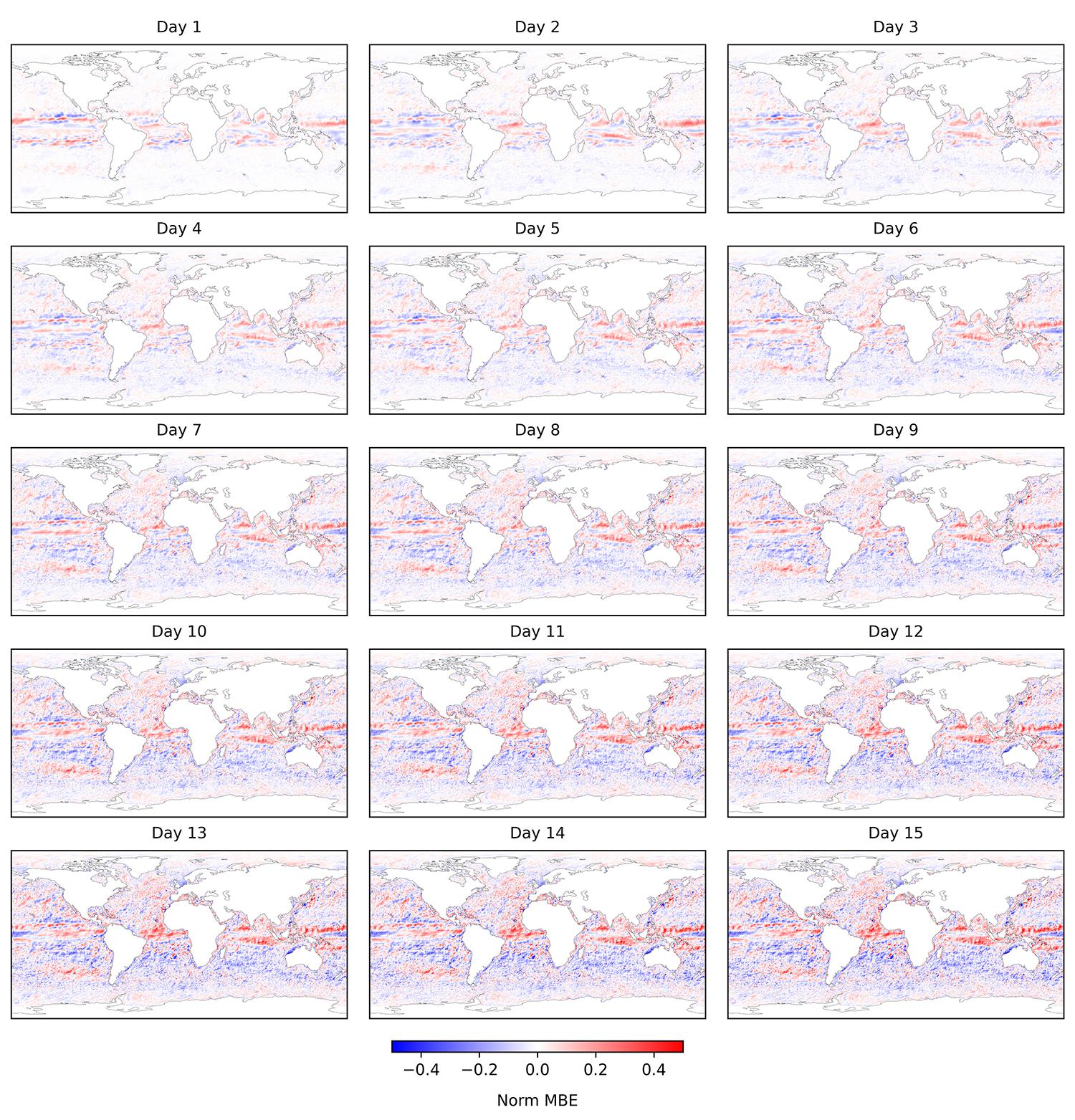}
	\caption{Normalized 15-day mean bias error (MBE) of physical fields (temperature, salinity, meridional current velocity, and zonal current velocity) in the KunPeng model, based on the 2021 test dataset.}
	\label{fig:E5}
\end{figure}

Fig.~\ref{fig:E5} illustrates the 15-day mean bias error of physical field predictions (temperature, salinity, meridional current velocity, and zonal current velocity) in the KunPeng model. The calculation formula is provided in Equation~\ref{eq:E1}.
\begin{equation}
	MBE^t_{hw} = \frac{1}{N_{var} \times N_{deep}} \sum_{c=1}^{N_{deep}} \sum_{v \in V} M_{chw} L_h \left( \hat{Y}^t_{cvhw} - Y^t_{cvhw} \right)
	\label{eq:E1}
\end{equation}
Let $V$ denote the physical field variable set and $N_{var}$ represent the cardinality of the variable set collection.

The mean deviation reflects the model's prediction tendencies across different regions during inference. Positive deviations (red) indicate the model tends to predict lower values compared to relatively higher true values, primarily concentrated in equatorial large-scale circulation centers, Northern Hemisphere regions, and Southern Hemisphere subtropical circulation centers. Negative deviations (blue) show the opposite pattern, with predictions higher than true values, mainly located at equatorial large-scale circulation boundaries and Southern Hemisphere regions. This suggests that while the model effectively reconstructs global-scale circulation, its predictions exhibit conservative biases. Introducing additional regularization constraints or region-specific loss weighting may improve performance.

\FloatBarrier

\subsection{Regional Prediction Results}

\begin{figure}[htbp]
	\centering
	\includegraphics[width=\linewidth]{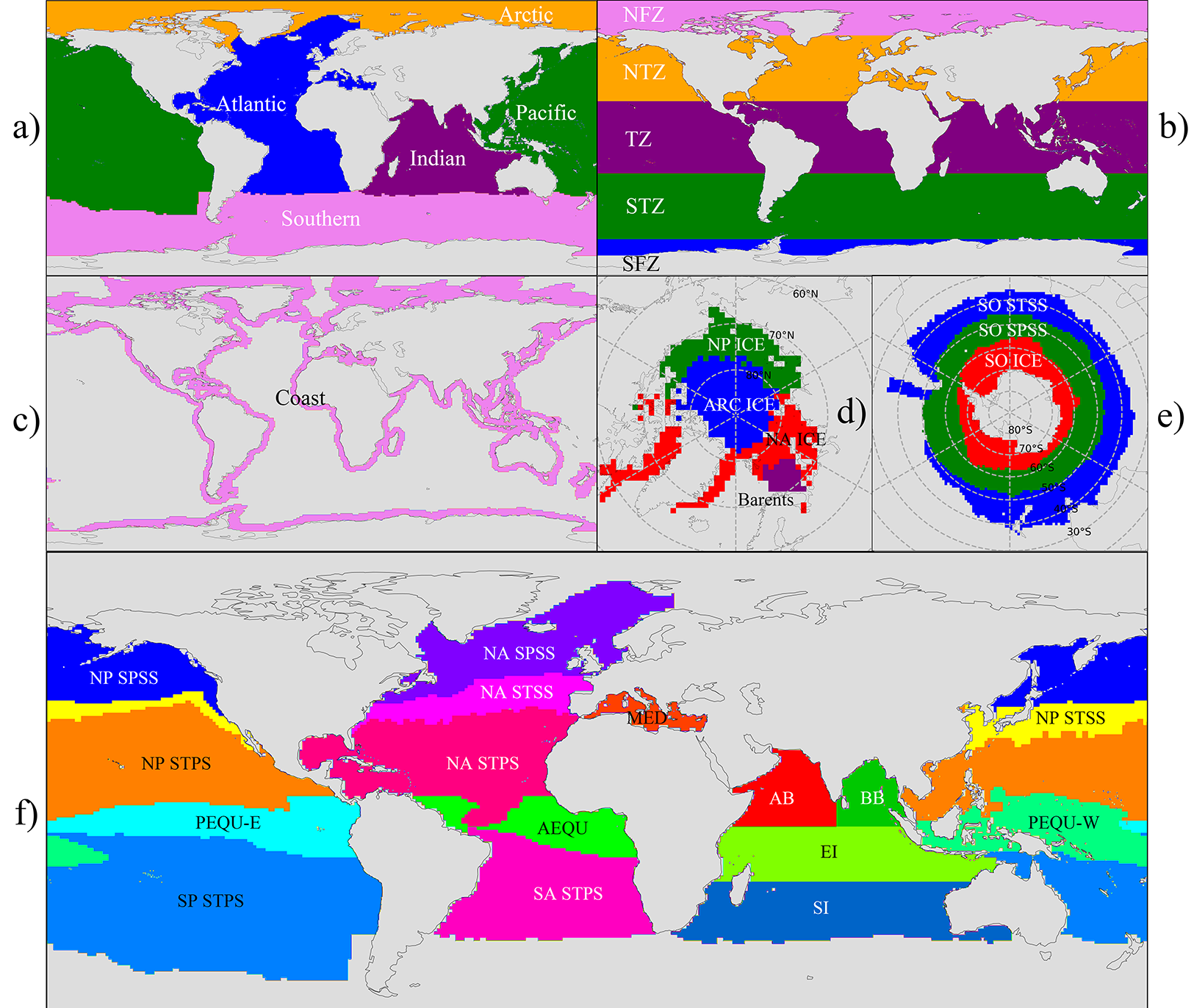}
	\caption{Schematic Classification of Global Oceans, Marine Zones, and Coastal Regions. \textbf{a)} Primary oceanic divisions: Pacific, Atlantic, Indian, Arctic, and Southern Oceans. \textbf{b)} Climate zone partitioning: North Frigid Zone (NFZ), North Temperate Zone (NTZ), Tropical Zone (TZ), South Temperate Zone (STZ), and South Frigid Zone (SFZ). \textbf{c)} Coastal regions adjacent to major continents, ice sheets, and island territories. \textbf{d–f)} Global marine zones classified according to the biome criteria of Fay and McKinley\cite{RN36}, including North Pacific Ice(NP ICE), Arctic Ice(ARC ICE), North Atlantic Ice(NA ICE), Southern Ocean Subtropical Seasonally Stratified (SO STSS), Southern Ocean Subpolar Seasonally Stratified (SO SPSS), Southern Ocean Ice(SO ICE), North Pacific Subpolar Seasonally Stratified(NP SPSS), North Pacific Subtropical Seasonally Stratified(NP STSS), North Pacific Subtropical Permanently Stratified(NP STPS), West pacific Equatorial(PEQU-W), East Pacific Equatorial(PEQU-E), South Pacific Subtropical Permanently Stratified(SP STPS), North Atlantic Subpolar Seasonally Stratified(NA SPSS), North Atlantic Subtropical Seasonally Stratified(NA STSS), North Atlantic Subtropical Permanently Stratified(NA STPS), Atlantic Equatorial(AEQU), South Atlantic Subtropical Permanently Stratified(SA STPS), Mediterranean(MED), Arabian Sea(AB), Bay of Bengal(BB), Equatorial Indian(EI), Southern Indian(SI)}
	\label{fig:E6}
\end{figure}

\begin{figure}[htbp]
	\centering
	\includegraphics[width=\linewidth]{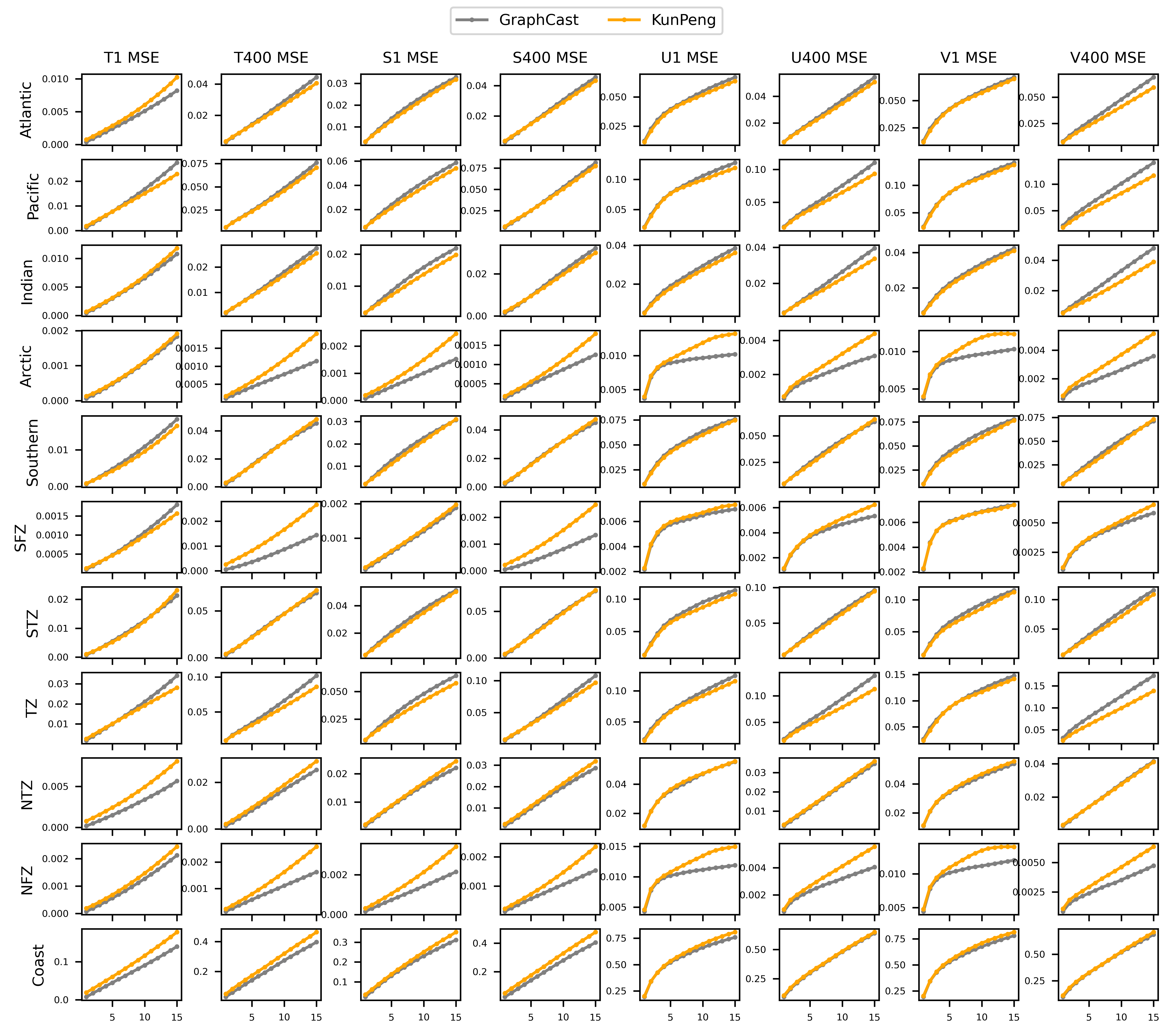}
	\caption{Comparison of MSE Metrics for 15-Day Forecasts between KunPeng and GraphCast in Major Oceans, Global Climate Zones, and Coastal Regions Worldwide}
	\label{fig:E7}
\end{figure}

\begin{figure}[htbp]
	\centering
	\includegraphics[width=0.8\linewidth]{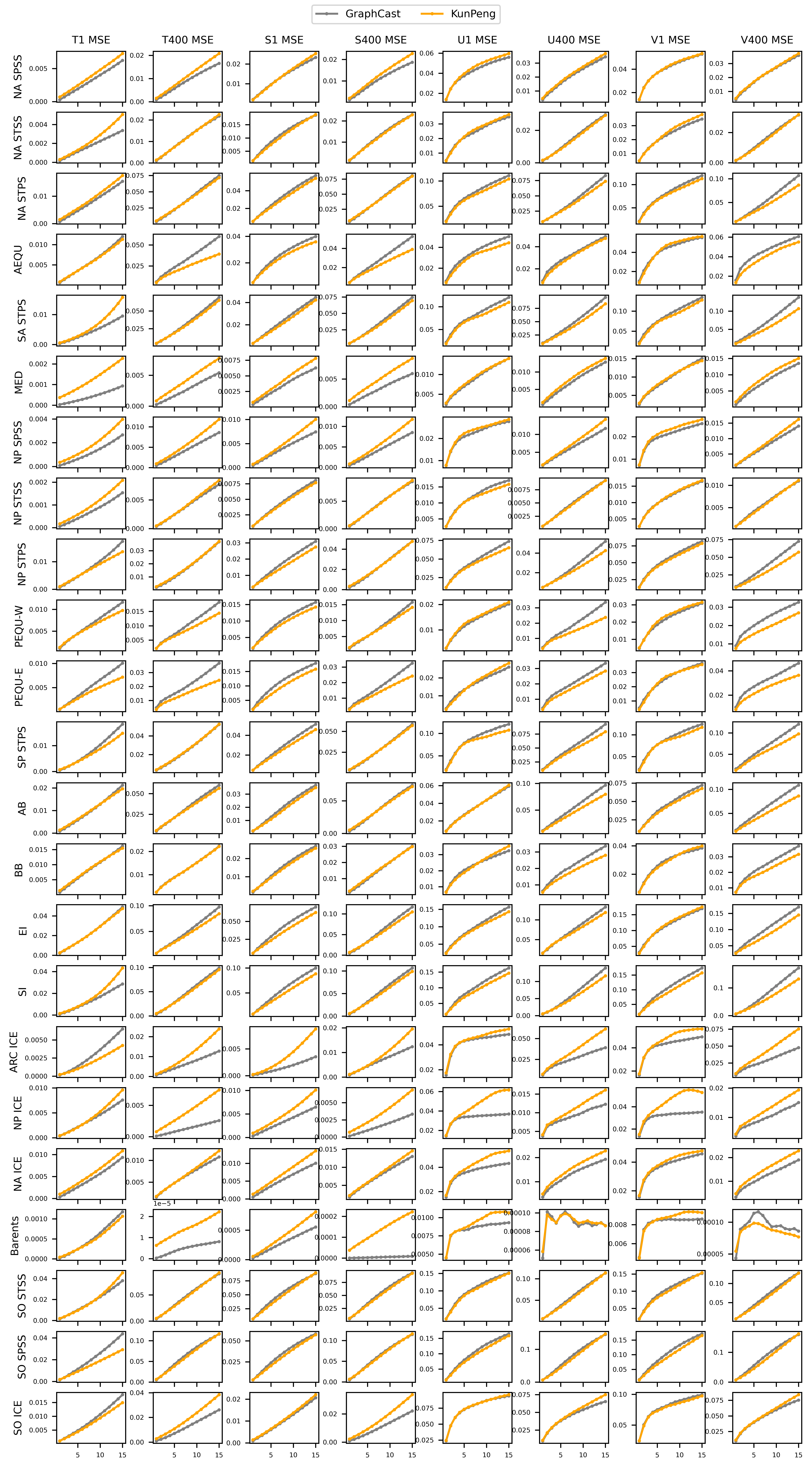}
	\caption{Comparison of MSE Metrics for 15-Day Forecasts between KunPeng and GraphCast in Major Oceanic Regions Worldwide}
	\label{fig:E8}
\end{figure}

The regional classification originates from the RECCAP2 project (REgional Carbon Cycle Assessment and Processes; \url{https://github.com/RECCAP2-ocean/R2-shared-resources}), which provides delineations of ocean basins, marine regions, and coastal zones represented as $1^\circ \times 1^\circ$ resolution masks (see Fig.~\ref{fig:E6}). The MSE results calculated for these regional divisions are presented in Figs.~\ref{fig:E7} and~\ref{fig:E8}.

The KunPeng model exhibits distinct geographical differences in prediction accuracy compared to the GraphCast baseline model. KunPeng outperforms GraphCast in most global regions for physical field metrics at both 1 m and 400 m ocean depths, with exceptions localized to polar and subpolar zones. Regions with relatively lower accuracy include: On the one hand, the polar regions are affected by seasonal sea ice, and the changes in physical field indicators present a complex trend of slow variation. It is necessary to introduce more environmental variables for reference, such as the thickness of polar ice layers, the sea ice area fraction, and so on. On the other hand, KunPeng is more likely to capture the changing laws of the physical field through the interaction of features at the spatial scale. However, for areas that are relatively closed in space, there is a lack of a way to reconstruct the physical field through the transfer of features at the temporal scale. It is worth noting that the depth of the Barents Sea ranges from 200 m to 400 m, with an average depth of 229 m. Therefore, the sample size of the sea layer at 400 m is relatively small, and the indicators do not have statistical significance.

\FloatBarrier

\subsection{Comparison Between KunPeng and GraphCast Baseline Models Across Metrics}

\begin{figure}[htbp]
	\centering
	\includegraphics[width=0.8\linewidth]{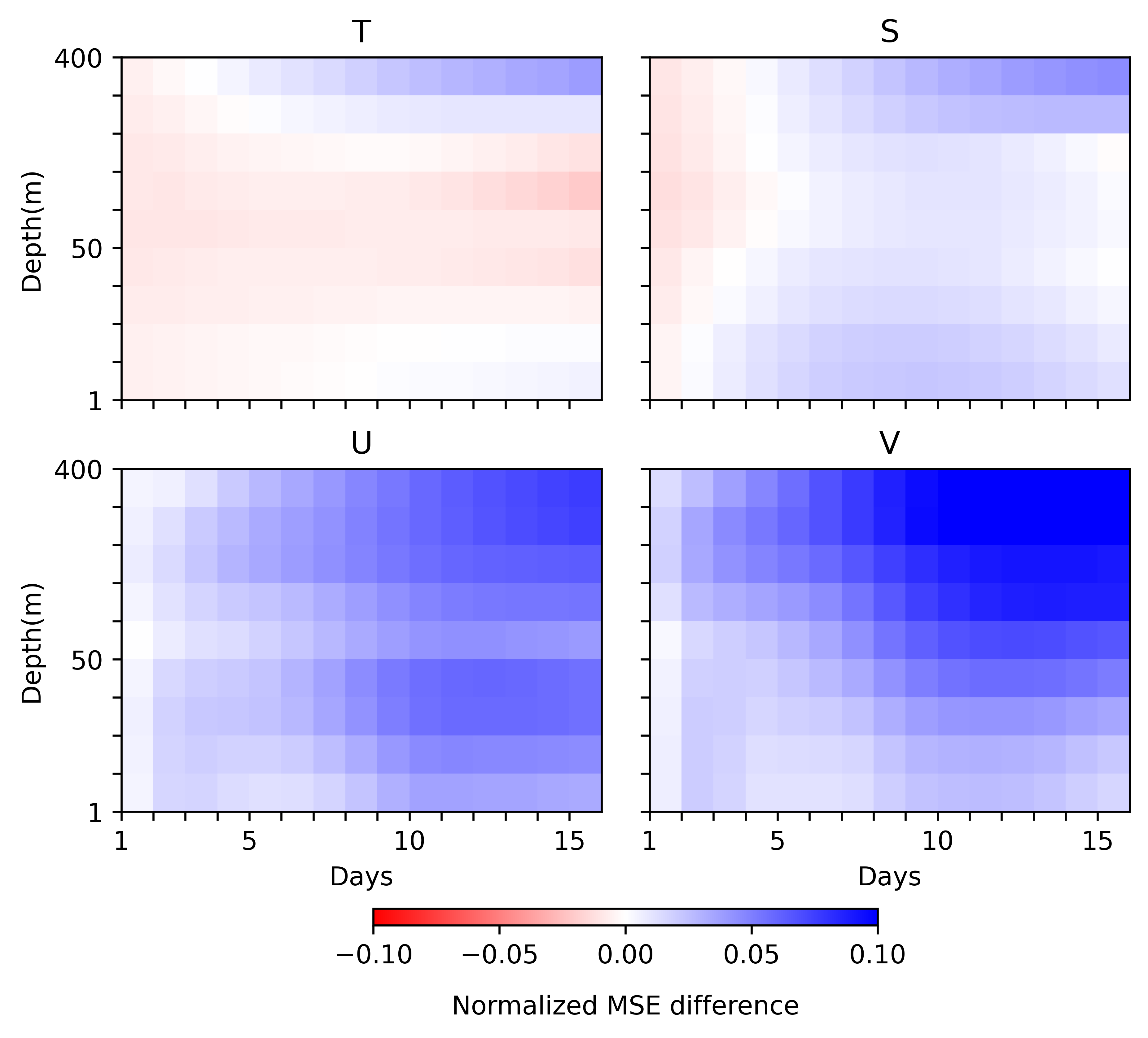}
	\caption{Normalized MSE Difference Maps for 15-Day Forecasts between KunPeng and GraphCast.Blue indicates regions where KunPeng outperforms GraphCast across specific time-depth-metric dimensions, while red denotes areas where KunPeng underperforms relative to GraphCast, based on the 2021 test set.}
	\label{fig:E9}
\end{figure}

\begin{figure}[htbp]
	\centering
	\includegraphics[width=\linewidth]{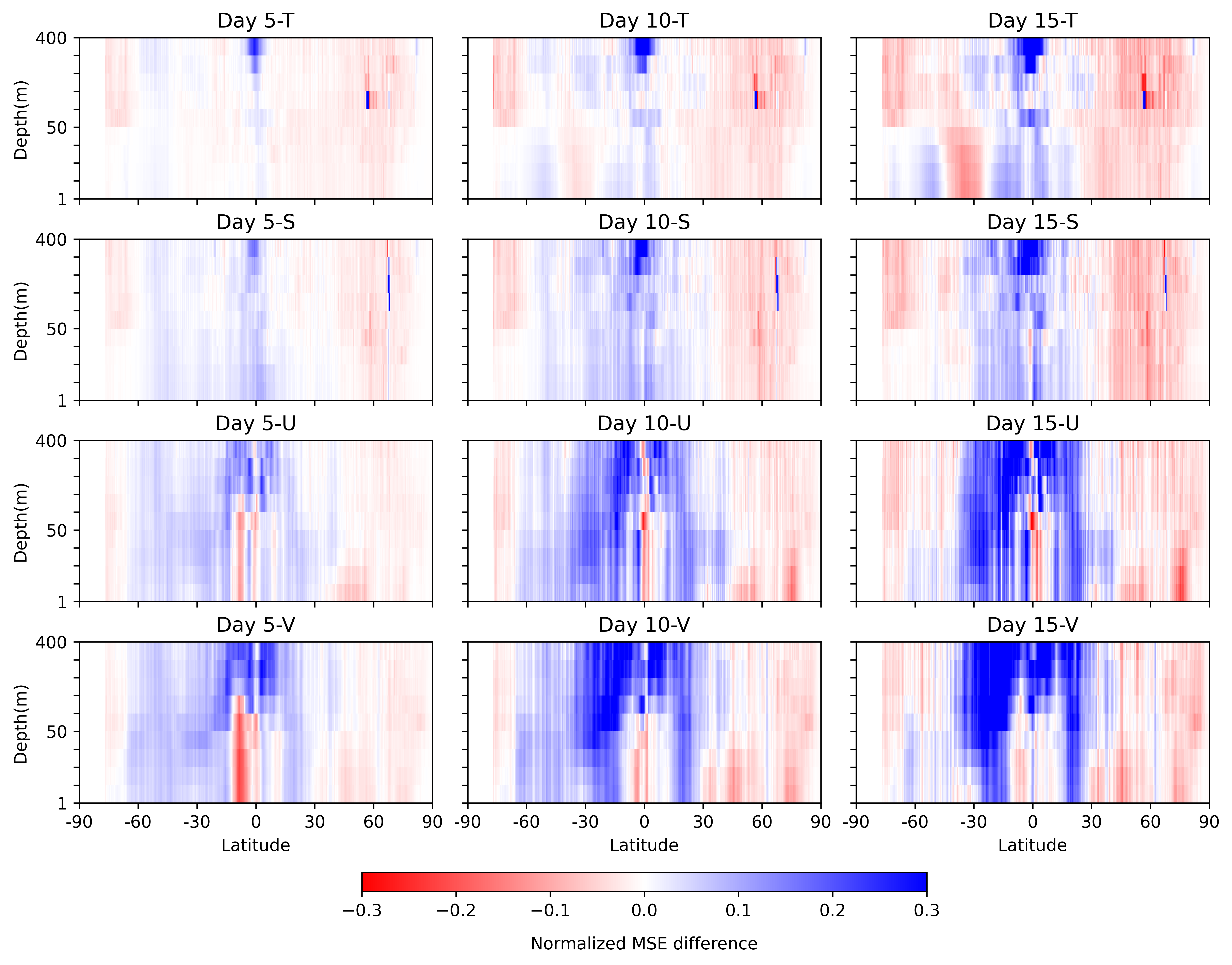}
	\caption{Latitude-Dependent Normalized MSE Difference Maps for 15-Day Forecasts between KunPeng and GraphCast.Blue indicates domains where KunPeng outperforms GraphCast across time-latitude-depth-metric dimensions, while red highlights regions where KunPeng underperforms relative to GraphCast, based on the 2021 test set.}
	\label{fig:E10}
\end{figure}

\begin{figure}[htbp]
	\centering
	\includegraphics[width=\linewidth]{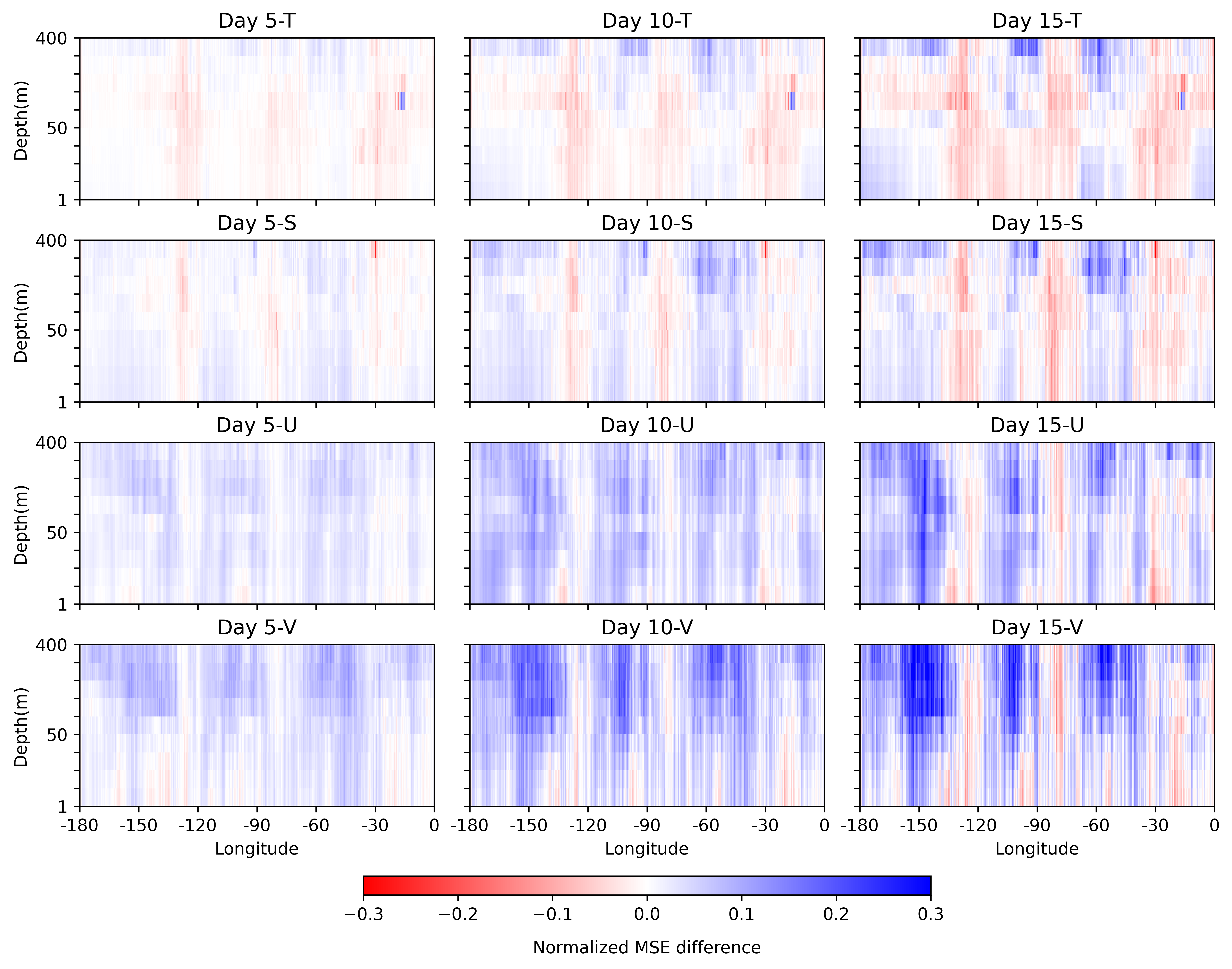}
	\caption{Longitude-Dependent Normalized MSE Difference Maps for 15-Day Forecasts between KunPeng and GraphCast. Blue indicates domains where KunPeng outperforms GraphCast across time-longitude-depth-metric dimensions, while red highlights regions where KunPeng underperforms relative to GraphCast, based on the 2021 test set.}
	\label{fig:E11}
\end{figure}

Figures~\ref{fig:E9}, \ref{fig:E10}, \ref{fig:E11} illustrate the normalized MSE differences in forecasting results between KunPeng and GraphCast, with statistics derived from the 15-day error averages spanning from \texttt{2021-01-01T00:00:00Z UTC} to \texttt{2021-12-16T00:00:00Z UTC}.
The comparative analysis reveals the following patterns: In temperature forecasting, GraphCast holds a relative advantage over KunPeng. For salinity predictions, GraphCast maintains lower MSE values in the first 5 days of prediction. However, KunPeng exhibits a significantly slower error accumulation rate, surpassing GraphCast in 5—15 day forecasts. Regarding velocity field modeling, KunPeng outperforms GraphCast across all metrics, with the performance gap widening as the forecast horizon extends.

\FloatBarrier

\section{Downstream Tasks}
\label{appendix:G}

The prediction accuracy of ocean physical fields by models can be evaluated through quantitative metrics. However, downstream prediction tasks based on ocean models require not only numerical forecasting accuracy but also consideration of additional physical constraints or multivariate interactions. To investigate model performance in ocean-related tasks and refine the global ocean model evaluation framework, we selected four representative tasks: mesoscale eddy prediction based on sea surface height to assess the smoothness of KunPeng's reconstructed ocean physical fields; mesoscale eddy prediction based on current velocity to evaluate KunPeng's capability in simulating dynamical processes of mesoscale phenomena; ocean front prediction based on sea surface temperature to test KunPeng's adaptability to regions with strong gradient variations; and fishing ground prediction for neon flying squid using ocean physical fields to assess KunPeng's comprehensive performance in nonlinear tasks.

The baseline ocean model employed was KunPeng. Under controlled experimental conditions with training epochs limited to 20, comparative results demonstrated KunPeng's superior performance. However, it should be noted that due to current computational resource limitations, we were unable to fully explore KunPeng's convergence limits - the loss function still has optimization potential, and the model's longitudinal performance requires further improvement.

\subsection{Mesoscale Eddies}

Mesoscale eddies are quasi-balanced vortex structures in the ocean with horizontal scales ranging from tens to hundreds of kilometers and temporal scales lasting from weeks to months. They serve as important carriers for energy transfer and material transport in the ocean. Their formation mechanisms are closely related to hydrodynamic instability, topographic effects, and wind stress forcing, and can be classified into two types: cyclonic (cold eddies) and anticyclonic (warm eddies). Through vertical pumping effects, mesoscale eddies significantly influence ocean thermal-salinity structures, vertical nutrient transport, and biogeochemical cycling processes. They can also carry water mass properties across thousands of kilometers, playing a crucial role in global ocean heat redistribution and climate regulation.

\begin{figure}[htbp]
	\centering
	\includegraphics[width=\linewidth]{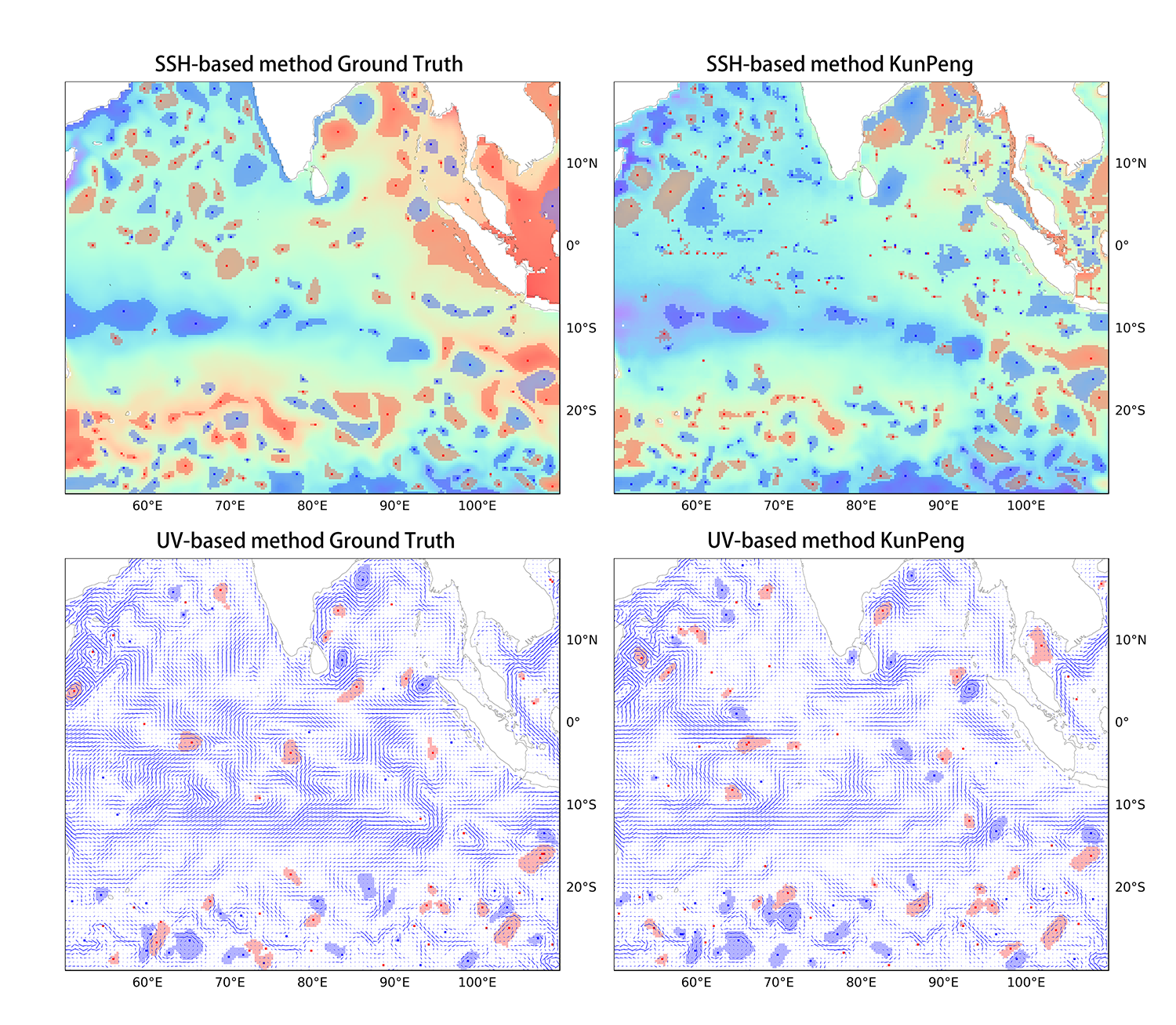}
	\caption{Mesoscale eddy prediction results. \(SSH\) denotes sea surface height, while \(U\) and \(V\) represent meridional and zonal current velocities, respectively. In the SSH-based identification method:Red and blue dots indicate cold (cyclonic) and warm (anticyclonic) eddy centers,and red and blue shaded areas represent cold and warm eddy regions.For the UV-based identification method:Red and blue dots denote clockwise and counterclockwise rotating eddies,and corresponding colored areas depict clockwise and counterclockwise eddy domains. It should be noted that clockwise rotating eddies manifest as warm-core (anticyclonic) vortices north of the equator, but as cold-core (cyclonic) vortices south of the equator, whereas counterclockwise eddies exhibit the opposite characteristics.}
	\label{fig:F1}
\end{figure}

The identification of mesoscale eddies relies on physical indicators that govern large-scale fluid motion, effectively reflecting whether model predictions adhere to physical constraints such as fluid dynamics and thermodynamics. Although there is currently no unified standard for determining the spatial extent of mesoscale eddies, any identification method can serve as a metric for evaluating the accuracy of model predictions in eddy detection. We employed two methods to identify eddy centers and their regions: one based on sea surface height (\(SSH\)) \cite{RN37} and another based on water velocity fields (\(U/V\)) \cite{RN38}. Figure~\ref{fig:F1} presents the identification results using both methods on analysis data from \texttt{2021-07-06T00:00:00Z UTC}, alongside the KunPeng model's 5-day predictions initialized from \texttt{2021-07-01T00:00:00Z UTC}. Both methods effectively identify mesoscale eddies. The SSH-based method locates eddy centers by detecting extrema in sea surface height and determines their spatial extent by progressively increasing threshold values with constant steps. While straightforward and capable of detecting numerous eddies, this method heavily depends on data accuracy and tends to generate significant noise in regions with gentle \(SSH\) gradients or small prediction differences, necessitating secondary filtering. In contrast, the UV-based method identifies eddy centers by analyzing rotational patterns in surrounding water flows and delineates eddy regions through path-integral contouring of streamlines relative to the eddy center. This approach offers higher reliability by excluding pseudo-eddies caused by large-scale circulations but requires strict contour adherence, making it more suitable for high-resolution data (e.g., 1 km $\times$ 1 km grids) and less adaptable to non-idealized eddy geometries in 0.25$^\circ$×0.25$^\circ$ resolution grids (27.8 km $\times$ 27.8 km at the equator).

\begin{figure}[htbp]
	\centering
	\includegraphics[width=\linewidth]{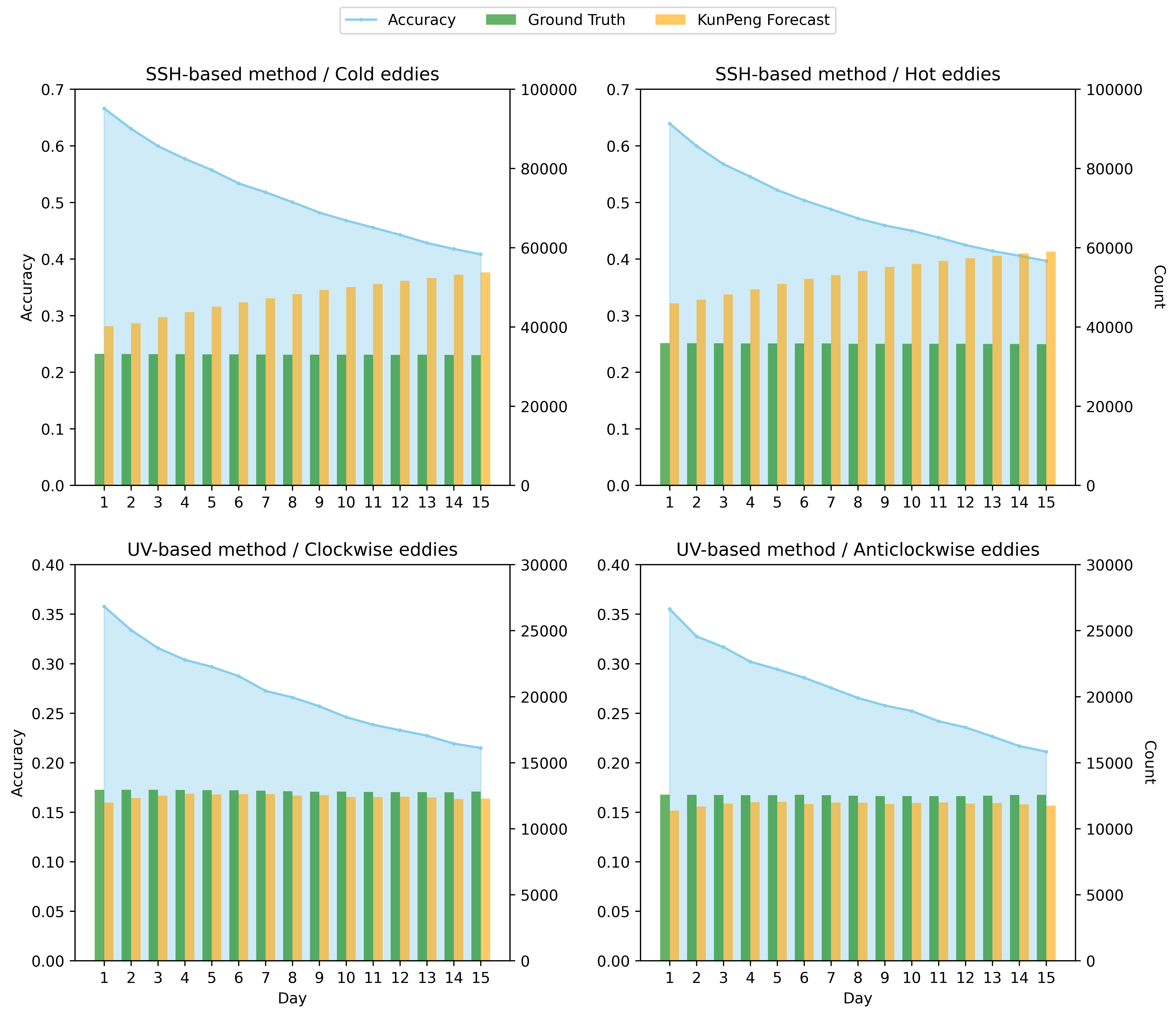}
	\caption{Statistical averages of mesoscale eddy identification. The prediction baseline period spans from \texttt{2021-07-01T00:00:00Z} to \texttt{2021-07-31T00:00:00Z UTC}, utilizing 15-day forecast results. The analysis employs two identification methods to calculate mesoscale eddy counts and coverage areas across the Pacific, Atlantic, and Indian Ocean basins.Accuracy represents the area prediction accuracy of eddy regions, while Count indicates the number of mesoscale eddies.}
	\label{fig:F2}
\end{figure}

To validate the model's accuracy in mesoscale eddy detection, we applied both identification methods to 15-day predicted physical fields initialized from July 2021 data, statistically evaluating eddy area accuracy and detection counts (see Figure ~\ref{fig:F2}). The SSH-based method achieved higher eddy area accuracy but produced significantly inflated eddy counts due to excessive noise from insufficiently smoothed predictions in regions with minor SSH variations. Conversely, the UV-based method yielded eddy center counts closely matching observations owing to the model's superior reconstruction of velocity fields, but its stringent morphological requirements and sensitivity to contour precision resulted in lower eddy area accuracy.

\FloatBarrier

\subsection{Ocean fronts}

\begin{figure}[htbp]
	\centering
	\includegraphics[width=\linewidth]{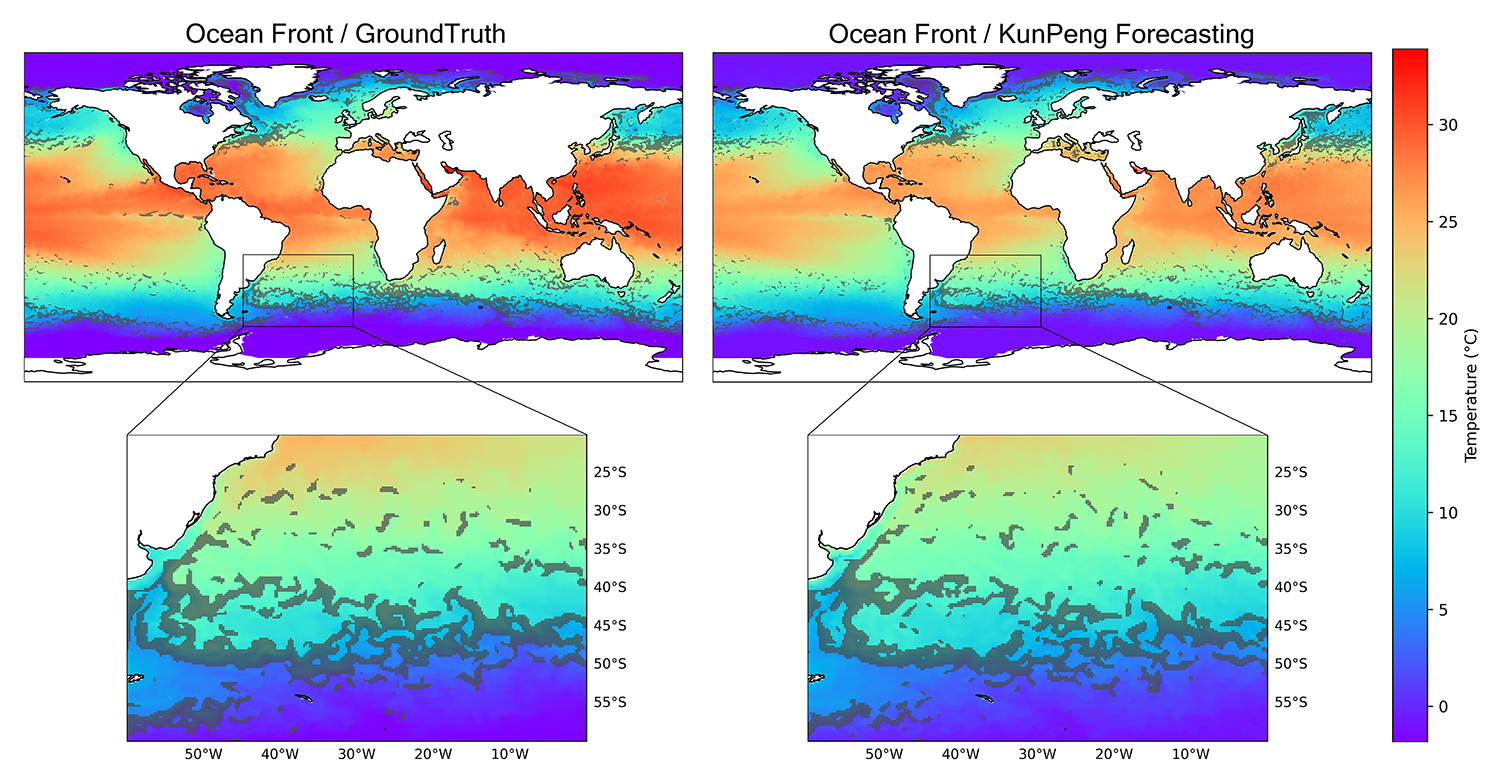}
	\caption{Comparative analysis of ocean thermal front predictions. The visualization displays results for \texttt{2021-07-11T00:00:00Z UTC}, with KunPeng model predictions generated from initial conditions at \texttt{2021-07-01T00:00:00Z UTC}. The heatmap shows the actual/predicted temperature. The gray areas are the calculated regions of the oceanic temperature fronts}
	\label{fig:F3}
\end{figure}

Ocean fronts are transitional zones between distinct water masses, characterized by sharp horizontal or vertical gradients in temperature, salinity, or density. These features span spatial scales from kilometers to hundreds of kilometers and temporal scales ranging from days to seasonal variations. As critical interfaces for energy transfer, material exchange, and biological activity, ocean fronts are not only fundamental to physical oceanography but also exert profound impacts on ecosystems and climate systems:Their intense vertical mixing drives upwelling of nutrient-rich deep waters, creating high-productivity fishing grounds,and they modulate heat and material exchange between the ocean and atmosphere, influencing regional-to-global climate patterns Ocean Thermal Fronts are identified through sea surface temperature (\(SST\)) gradient calculations \cite{RN39}, as defined by Equation ~\ref{eq:sst_grad}:

\begin{equation}
	|\Delta SST|(w,h) = \sqrt{\left(\frac{\partial SST}{\partial x}\right)^2 + \left(\frac{\partial SST}{\partial y}\right)^2}
	\label{eq:sst_grad}
\end{equation}

where $SST$ represents sea surface temperature, while $x$ and $y$ denote the longitudinal and latitudinal distances, respectively. The temperature partial derivatives at each grid point are computed using the centered difference method, expressed as:

\begin{equation}
	\frac{\partial SST}{\partial x}(h,w) = \frac{SST(h, w + 1) - SST(h, w - 1)}{2\Delta x(w)}
	\label{eq:sst_grad_x}
\end{equation}

\begin{equation}
	\frac{\partial SST}{\partial y}(h,w) = \frac{SST(h + 1, w) - SST(h - 1, w)}{2\Delta y}
	\label{eq:sst_grad_y}
\end{equation}

where $\Delta y$ represents the latitudinal grid spacing at the given resolution (27.8 km for 0.25$^\circ$ $\times$0.25$^\circ$ grids), $\Delta x(w)$ denotes the longitudinal grid spacing (27.8$\times \cos(\phi)$ km for 0.25$^\circ$ $\times$0.25$^\circ$ grids), and $\phi$ is the latitude value. After computing temperature gradients, the upper quintile of gradient magnitudes identifies ocean thermal front regions. Figure ~\ref{fig:F3} presents a comparison between the observed global ocean frontal values and the 10-day forecasted results computed by the KunPeng model.
To evaluate the KunPeng model's performance in ocean front detection, we employed two quantitative metrics: Intersection over Union (IoU) and F1-score, with their mathematical formulations provided in Equations ~\ref{eq:iou} and ~\ref{eq:f1} respectively.

\begin{equation}
	IoU^{l} = \frac{\sum_{t=1}^{T}S_{truth}\cap S^{l}_{pred}}{\sum_{t=1}^{T}S_{truth} \cup S^{l}_{pred}}
	\label{eq:iou}
\end{equation}

\begin{equation}
	F1 = \frac{2 \times Precision \times Recall}{Precision + Recall}
	\label{eq:f1}
\end{equation}

\begin{figure}[htbp]
	\centering
	\includegraphics[width=\linewidth]{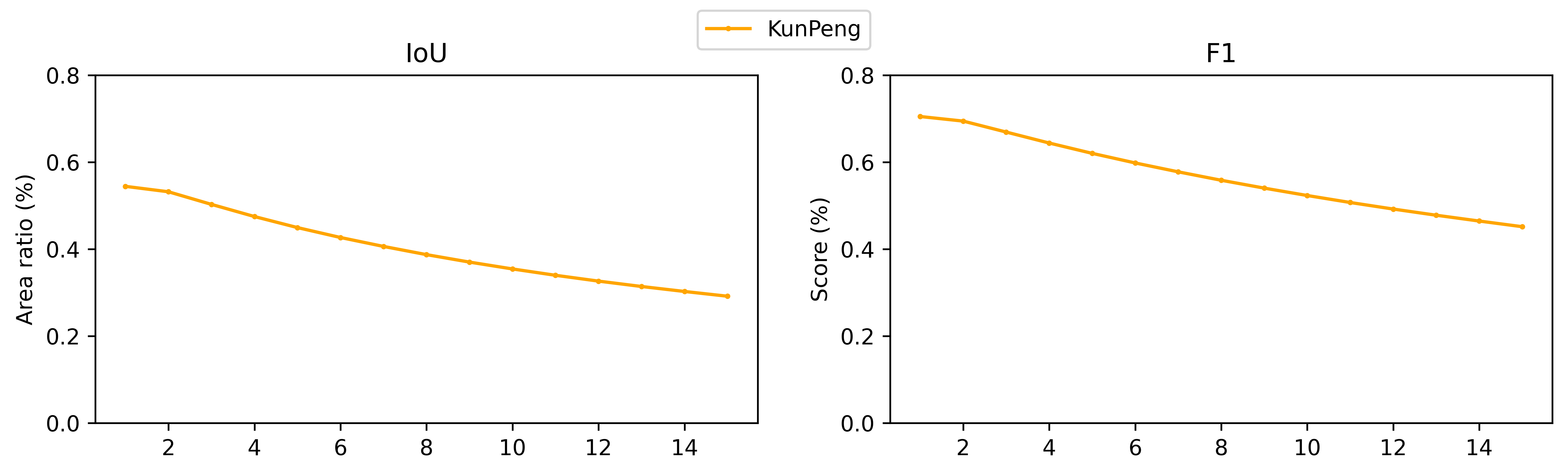}
	\caption{Statistical analysis of ocean thermal fronts based on KunPeng model predictions. The prediction baseline period spans from \texttt{2021-01-01T00:00:00Z} to \texttt{2021-12-16T00:00:00Z UTC}, utilizing 15-day forecast results with global oceanic coverage}
	\label{fig:F4}
\end{figure}

Where $S_{truth}$ represents the actual area of the oceanic front, $S_{pred}$ represents the predicted area of the oceanic front, $l$ represents the prediction step size, $Precision$ represents the precision rate, which is the proportion of correctly predicted positive samples among all predicted positive samples, and $Recall$ represents the recall rate, which is the proportion of correctly predicted positive samples among all actual positive samples. Statistical results are shown in Figure ~\ref{fig:F4}.

Unlike the mesoscale eddies with smooth expectations, the identification of oceanic fronts requires adaptation to the rapid changes in the gradients of variables in the oceanic thermohaline field, especially the mesoscale effects generated by the interactions in the adjacent areas of different sea regions. In the experiment, we effectively identified the regions with gradient changes higher than 0.014$^\circ$C/km through the upper 5th percentile, adapted to the short-term changes of the fronts at the oceanic confluence areas, and retained the long-term influence of the planetary-scale fronts, which proves that the KunPeng model has a high degree of accuracy in predicting oceanic mesoscale phenomena.

\FloatBarrier

\subsection{Dosidicus gigas fishery}

The Humboldt squid (Dosidicus gigas) represents a commercially vital species in pelagic fisheries. This cephalopod exhibits a broad distribution across tropical to temperate waters of the eastern Pacific Ocean and the Indian Ocean, characterized by rapid growth rates, a short life cycle (approximately 1-2 years), and diel vertical migration behavior. As an apex predator, it plays a pivotal role in marine ecosystems while serving as a primary target for distant-water fisheries of multiple nations due to its high productivity and commercial value. In recent years, climate change and oceanic variability have driven significant interannual fluctuations in its abundance and spatial distribution, necessitating the development of environmentally-driven predictive models for accurate fishing ground forecasting to support sustainable resource utilization and fisheries management decisions.

\begin{figure}[htbp]
	\centering
	\includegraphics[width=\linewidth]{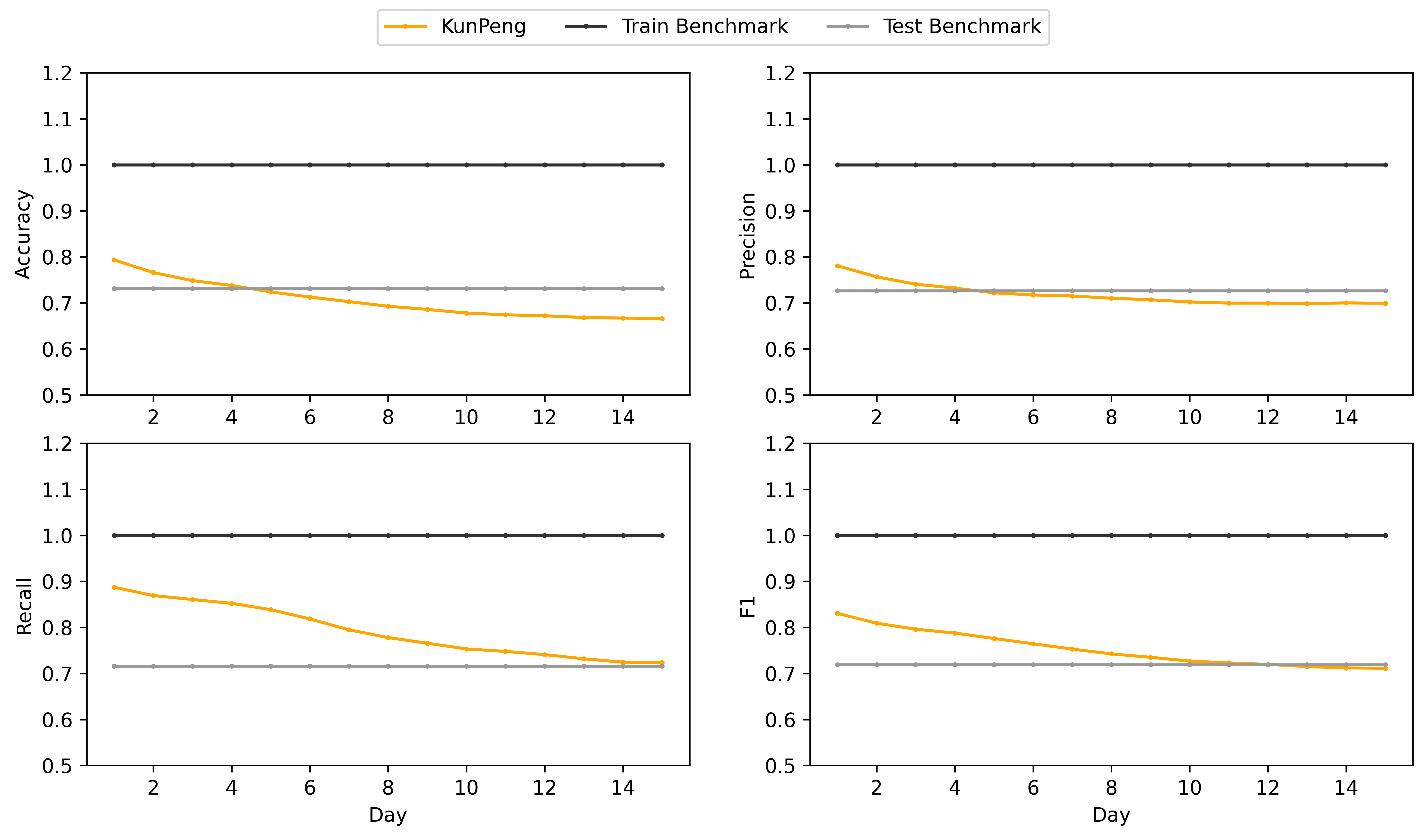}
	\caption{Prediction indices of the fishing grounds for Dosidicus gigas based on the predicted values of the KunPeng model. The basic prediction dates range from \texttt{2021-01-01T00:00:00Z UTC} to \texttt{2021-12-16T00:00:00Z UTC}. The 15-day prediction results are used, and the prediction area is the global sea area. The Train Benchmark is the index calculated for the training set of the fishing grounds for Dosidicus gigas when the XGBoost model uses the actual values of ocean variables. The Test Benchmark is the index obtained by using the test set. KunPeng calculates the fishing grounds by using the environmental prediction variables of all the dates in both the training set and the test set}
	\label{fig:F5}
\end{figure}

To evaluate the predictive accuracy of the KunPeng ocean model for Dosidicus gigas fishing grounds based on temporally reconstructed physical ocean fields, we utilized 2021 squid fishery data from National Data Centre for Distant-water Fisheries of China. The prediction was performed using XGBoost algorithm incorporating six key oceanic parameters: temperature, salinity, zonal (\(U\)) and meridional (\(V\)) current velocities, mixed layer depth, and sea surface height. Fishing grounds were classified using a CPUE (Catch Per Unit Effort) threshold of 3.0, with areas $\geq$ 3.0 defined as fishing grounds and areas $<$ 3.0 as non-fishing grounds. CPUE calculation detailed in Equation ~\ref{eq:cpue}.

\begin{equation}
	CPUE_{h,w} = \frac{Catch_{h,w}}{Effort_{h,w}}
	\label{eq:cpue}
\end{equation}

where $Catch$ represents the total catch (in metric tons) within each grid cell, while $Effort$ denotes the corresponding fishing effort (number of hauls). During baseline development, the annual dataset was randomly partitioned into 80\% training data and 20\% testing data to train an XGBoost-based fishing ground prediction model using in situ environmental observations. For model validation, 15-day environmental forecasts generated by the KunPeng model (spanning \texttt{2021-01-01T00:00:00Z} to \texttt{2021-12-16T00:00:00Z UTC}) were employed to predict fishing ground distributions. Model performance was quantified through four metrics: accuracy, precision, recall, and F1-score, with detailed results presented in Figure ~\ref{fig:F5}.

Unlike physical quantities and physically derived indices, fishing ground prediction requires that the prediction process fully takes into account relevant environmental factors, establishes nonlinear dependency relationships, and simultaneously incorporates prior knowledge of various fish species. For example, studies have shown that as a deep-sea cephalopod with vertical migration behavior, the Dosidicus gigas is significantly influenced by six factors: sea surface height, sea water temperature at a depth of 50 m, mixed layer depth, surface sea water temperature, surface salinity, and sea water temperature at a depth of 250 m \cite{RN40}. Therefore, the model needs to learn the correlations among different oceanographic indices to improve its application prospects and scope of adaptability. Experiments have shown that the prediction of fishing grounds for the Humboldt squid based on the ocean physical fields reconstructed by the KunPeng model has relatively high accuracy and can be applied to prediction scenarios related to distant-water fishing in the fishery industry

\FloatBarrier

\end{document}